\pdfoutput=1

\documentclass[11pt]{article}

\usepackage[]{acl}

\usepackage{times}
\usepackage{latexsym}

\usepackage[T1]{fontenc}

\usepackage[utf8]{inputenc}

\usepackage{amsmath}
\usepackage{graphicx}
\usepackage{tabularx}
\usepackage{url}
\usepackage{hyperref}
\usepackage{framed}
\usepackage{booktabs}
\usepackage{multirow}
\usepackage{lipsum} 
\usepackage{color, colortbl}
\usepackage{pbox}
\usepackage{adjustbox}

\newcommand{\mask}[0]{\rule{.8cm}{0.15mm}}

\usepackage[ruled, vlined]{algorithm2e}
\usepackage{multirow}
\usepackage{amsfonts}
\usepackage{bbm}

\usepackage{microtype}

\title{How to Query Language Models?}

\author{
  Leonard Adolphs \\
  \And
  Shehzaad Dhuliawala \\
  \\
  Department of Computer Science \\
  ETH Zürich \\
  \texttt{\{\textit{firstname}.\textit{lastname}\}@inf.ethz.ch} \\\And
  Thomas Hofmann \\
  }

\begin{document}
\maketitle

\begin{abstract}
Large pre-trained language models (LMs) are capable of not only recovering linguistic but also factual and commonsense knowledge.
To access the knowledge stored in mask-based LMs, we can use cloze-style questions and let the model fill in the blank. 
The flexibility advantage over structured knowledge bases comes with the drawback of finding the right query for a certain information need. Inspired by human behavior to disambiguate a question, we propose to query LMs by example. To clarify the ambivalent question \textit{Who does Neuer play for?}, a successful strategy is to demonstrate the relation using another subject, e.g., \textit{Ronaldo plays for Portugal. Who does Neuer play for?}. We apply this approach of querying by example to the LAMA probe and obtain substantial improvements of up to 37.8\% for BERT-large on the T-REx data when providing only 10 demonstrations---even outperforming a baseline that queries the model with up to 40 paraphrases of the question. The examples are provided through the model's context and thus require neither fine-tuning nor an additional forward pass. This suggests that LMs contain more factual and commonsense knowledge than previously assumed---if we query the model in the right way.

\end{abstract}
\section{Introduction}
Language Models (LM) are omnipresent in modern NLP systems. In just a few years, they've been established as the standard \textit{feature extractor} for many different language understanding tasks \citep{karpukhin2020dense, zhang2020retrospective, wang2019structbert, he2020deberta}. Typically, they are used to create a latent representation of natural language input and then fine-tuned to the task at hand. However, recent work \citep{lama, whatlmknow, gpt3, 2020arXiv200208910R} has shown that \textit{off-the-shelve} language models capture not only linguistic features but also large amounts of relational knowledge, not requiring any form of re-training. 
\begin{figure}[t]
    \centering
     \includegraphics[width=.7\columnwidth]{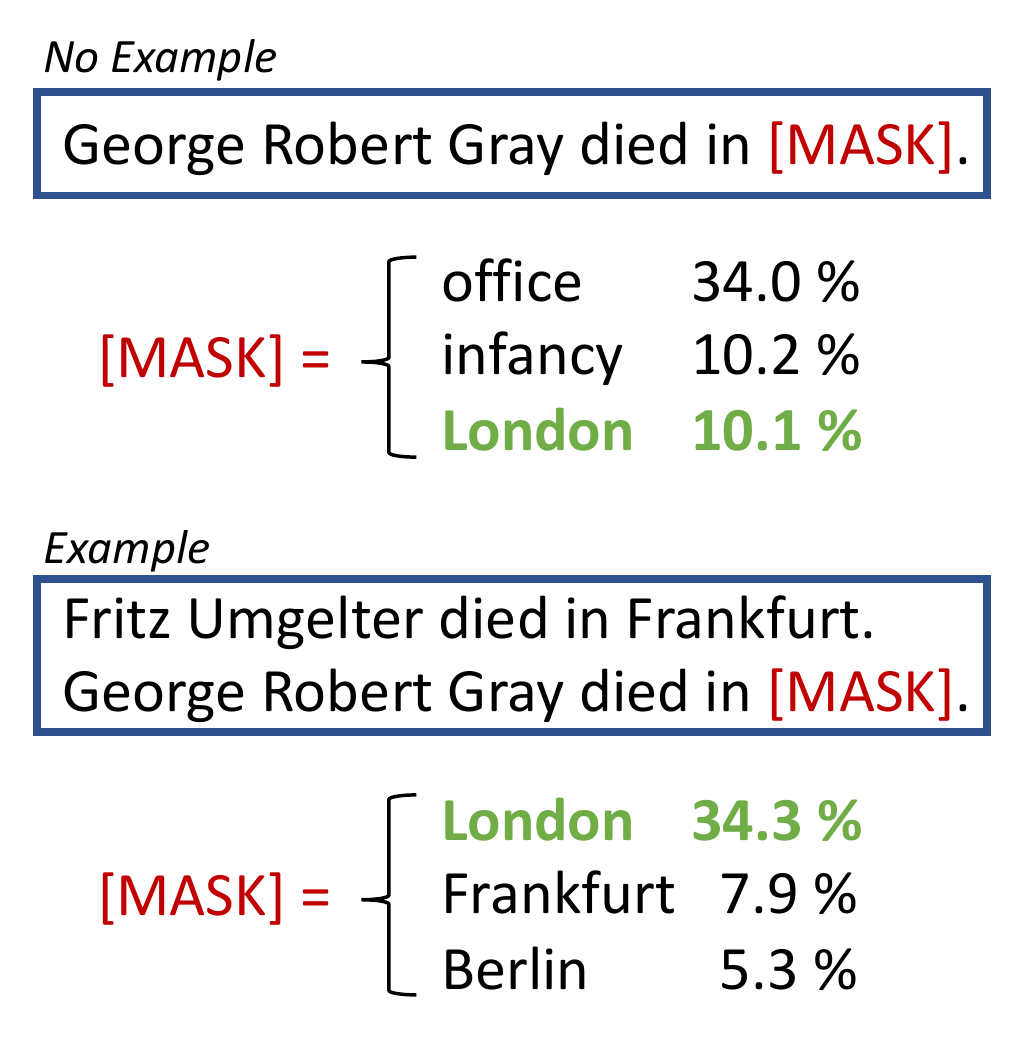}  
    \caption{BERT's top-3 predictions with probabilites when prompted with the cloze-style question (top) versus when prompted with one additional example of the same relation (bottom).}
    \label{fig:overview}
\end{figure}

The LAMA probe by \citet{lama} was designed to quantify the amount of relational knowledge present in (mask-based) language models. While the task of predicting the right object for a subject-relation tuple remains the same as for a standard knowledge base (KB) completion query, the input is structured in a cloze-style sentence. For example, a KB completion query of the form \textit{(Dante, born-in, X)} becomes \textit{"Dante was born in [MASK]."}. \citet{lama} show that BERT~\citep{devlin2019bert} performs on par with competitive specialized models on factual and commonsense knowledge. The performance on this task can only be seen as a lower bound to the actual knowledge present in language models as the choice of natural language template for a given relation might be suboptimal \citep{lama, whatlmknow}. The more general question here is \textit{"How to query an LM for a specific information need?"}. \citet{whatlmknow} propose to use multiple paraphrases of the probe and then aggregate the solutions. \citet{petroni2020context}, on the other hand, add relevant context. Both approaches can be linked to common human behavior. In human dialog, a question can be made more precise both by paraphrasing or adding additional context information. Since language models are trained on large amounts of human-generated data, the intuition of phrasing the information need \textit{most naturally} seems obvious. Humans excel at pattern recognition and pattern continuation for many different modes of representation \citep{doi:10.1007/BF02833890}. Concepts embedded in language are no exception to this. Therefore, another common way to probe a human's knowledge is by providing examples and asking them to transfer the relation provided to a new object. For example, asking \textit{Who plays Neuer for?} is ambiguous as both \textit{Bayern Munich} and \textit{Germany} would be correct answers. However, when contextualizing the question with an example, the answer is clear: \textit{I know Ronaldo plays for Portugal. Who plays Neuer for?}.

In this work, we apply the concept of querying by example to probe language models. Additional to the cloze-style question, we provide other examples of the same relation to the model's input. The previous example's input then becomes \textit{"Ronaldo plays for Portugal. Neuer plays for [MASK]."}. We show that by providing only a few demonstrations, standard language models' prediction performance improves drastically. So much so that for the TREx dataset, it becomes an even more powerful technique to retrieve knowledge than using an ensemble of up to 40 different paraphrases~\citep{whatlmknow}, while requiring only a single forward pass instead of 40. 
\section{Related Work}

\paragraph{Language Model Probes} 
\citet{lama} started to investigate how much factual and commonsense knowledge LMs posses. They released the LAMA probe, which is a dataset consisting of T-REx \citep{trex}, Google-RE, ConceptNet \citep{speer2018conceptnet}, and SQuAD \citep{DBLP:journals/corr/RajpurkarZLL16}. Each dataset is transformed to be a collection of $\langle$subject, relation, object$\rangle$-triplets and pruned to only contain single token objects present in BERT's vocabulary. Additionally, they provide templates in natural language for each relation. Their investigation reveals that BERT-large has remarkable capabilities in recalling factual knowledge, competitive to supervised baseline systems. \\
Since there is usually more than one way to express a relation, the LAMA probe score can only be regarded as a \emph{lower bound} \citep{lama, whatlmknow}. To tighten this lower bound, \citet{whatlmknow} propose an automatic discovering mechanism for paraphrases together with an aggregation scheme. By querying the LM with a diverse set of prompts, they significantly improve the LAMA probe's baseline numbers for BERT models. However, this approach incurs the cost of additional queries to the LM, an optimization procedure to aggregate the results, and the extraction of paraphrases.\\
Machine reading comprehension (MRC) and open-domain question answering (QA) are fields in NLP dominated by large pre-trained LMs. Here, the premise typically is that the model is capable of extracting the answer from the provided context, rather than having it stored in its parameters\footnote{With the notable exception of the work of \citet{2020arXiv200208910R}, which uses a T-5 model without any access to an additional knowledge base.}. \citet{petroni2020context} extend this line of thought to retrieve factual knowledge from LMs by providing relevant context but \emph{without} fine-tuning the model. Their experiments show that providing relevant passages significantly improves the scores on the LAMA probe for BERT models.

\paragraph{Few-Shot Learning} The term few-shot learning refers to the practice of only providing a few examples when training a model, compared to the typical approach of using large datasets \citep{wang2020generalizing}. In the NLP domain, recent work by \citet{gpt3} suggests to use these few examples only in the context, as opposed to actually \emph{training} with it. Fittingly, they call this approach \textit{in-context} learning. Here, they condition the model on a natural language description of the task together with a few demonstrations. Their experiments reveal that the larger the model, the better its in-context learning capabilities. Our approach is very similar to in-context learning, with the difference that we do not provide a description of the task and utilize natural language templates for the relations. The motivation is that this should closely resemble human behavior of providing examples of a relation: instead of providing a list of subject and objects and let the other person figure out the relation, a human typically provides the subject and objects embedded in the template relation. Moreover, we understand our approach not as a \emph{learning} method, but rather as a querying technique that disambiguates the information need. \\
\citet{smalllmfewshot} argue that small LMs can be effective for few-shot learning too. However, they approach the problem of limited examples differently; instead of providing it as conditioning in the input, they actually train with it. By embedding the data into relation templates, they obtain training data that is closer in style to the pre-training data and, thus, can learn with fewer samples. \citet{gao2020making} take this concept even further and automate the template generation. Additionally, they also find that---when fine-tuning with few samples---providing good demonstrations in the context improves the model's performance.
\section{Background}

\subsection{Language Models for cloze-style QA}
In this work, we probe mask-based language models for their relational knowledge. The considered facts are triplets consisting of a subject, a relation, and an object $\langle s, r, o \rangle$. Language models are trained to predict the most probable word given the (surrounding) context. Hence, to test a model's factual knowledge, we feed it natural text with the object masked out. This requires a mapping from the relation $r$ to a natural language prompt $t_r$ with placeholders for subject and object, e.g., the relation $r$ = \textit{age} becomes $t_r$ = \textit{[s] is [o] years old}.
When probing for a single $\langle s, r, o \rangle$-triplet, the input to the language model is the natural language prompt $t_r$ of the relation $r$ together with the subject $s$. It outputs a \textit{likelihood} score $\text{P}_{\text{LM}}$ for each token in its vocabulary $\mathcal{V}$ which we use to construct a top-$k$ prediction subset $\mathcal{V}'$ for the object $o$:
\begin{align}
    \mathcal{V}' = \arg \max_{\mathcal{V}' \subset \mathcal{V}, |\mathcal{V'}| = k} \sum_{o' \in \mathcal{V}'} \text{P}_{\text{LM}}(o' | s, t_r)
\end{align}
The language model \textit{succeeds} for the triplet @$k$ if $o \in \mathcal{V'}$. For example, we say that it knows the fact $\langle s=$ Tiger Woods$, r= $ age$, o=45 \rangle$ @3, if for the query \textit{"Tiger Woods is [MASK] years old"} it ranks the token \textit{"45"} within the top-3 of the vocabulary.


\subsection{Datasets}

\begin{table}[h]
\centering
\small
    \begin{tabular}{llcc}
    \toprule
    \multirow{2}{*}{Corpus} & \multirow{2}{*}{Relation} & \multicolumn{2}{c}{Statistics} \\
             & & \#Facts & \#Relations \\
    \midrule
    \multirow{4}{*}{Google-RE}  & birth-place & 2937 & 1  \\
    &birth-date & 1825 & 1 \\
    &death-place & 765 & 1 \\
    \cmidrule{2-4}
    & Total & 5527 & 3 \\
    \midrule
    \multirow{4}{*}{T-REx}  & $1$-$1$ & 937 &2 \\
    &$N$-$1$ & 20006 & 23 \\
    & $N$-$M$ & 13096 & 16 \\
    \cmidrule{2-4} 
    & Total & 34039 & 41 \\
    \midrule 
    ConceptNet & Total & 11458 & 16 \\
    \bottomrule
    \end{tabular}

\caption{Statistics for the corpora of the LAMA data.}
\label{tab:data}
\end{table}

We use the LAMA probe in our experiments~\citep{lama}. It's a collection of factual and commonsense examples provided as $\langle s, r, o \rangle$-triplets\footnote{We do not consider the SQuAD dataset of the probe as it has no clear notion of \textit{relation}.} with single token objects. Moreover, it provides human-generated templates $t_r$ for each relation $r$.
The statistics about the three considered corpora T-REx~\citep{trex}, Google-RE\footnote{\tiny{\url{https://github.com/google-research-datasets/relation-extraction-corpus}}}, and ConceptNet \citep{speer2018conceptnet} are provided in Table~\ref{tab:data}.

\subsection{Models}
We investigate the usefulness of querying by example, for three individual language models: BERT-base, BERT-large \citep{devlin2019bert}, and ALBERT-xxl \citep{lan2020albert}. These models are among the most frequently used language models these days\footnote{According to the statistics from \url{https://huggingface.co/models?filter=pytorch,masked-lm}.}.
For both BERT models, we consider the cased variant, unless explicitly noted otherwise.
\section{Method}
\label{sec:method}
Our proposed method for querying relational knowledge from LMs is simple yet effective. When we construct the query for the triplet $\langle s, r, o\rangle$, we provide the model with additional samples $\{\langle s', r, o'\rangle, \langle s'', r, o''\rangle, \dots \}$ of the same relation $r$. These additional examples are converted to their natural language equivalent using the template $t_r$ and prepend to the cloze-style sentence representation of $\langle s, r, o\rangle$. The intuition is that the non-masked examples provide the model with an idea of filling in the gap for the relation at hand. As can be seen in Figure~\ref{fig:overview}, providing a single example in the same structure clarifies the object requested for both humans and BERT. This is particularly useful when the template $t_r$ does not capture the desired relation $r$ between subject $s$ and object $o$ unambiguously, which in natural language is likely to be the case for many relations. In this sense, it tries to solve the same problem as paraphrasing. A query is paraphrased multiple times to align the model's understanding of the query with the actual information need. When we provide additional examples, we do the same by showing the model how to apply the relation to other instances and ask it to generalize. Of course, the model does not reason in this exact way; rather, through its training data, it is biased towards \emph{completing patterns} as this is a ubiquitous behavior in human writing.\\

\begin{table}[h]
\centering
\small
    \begin{tabular}{p{4.1cm}p{2.9cm}}
    \toprule
    Query & Predictions \\
    \midrule
    \textbf{No Example} & \\
    \hspace{.1cm} Rodmarton\footnotemark is a \mask . & farmer \small{(3.9\%)} \\ & businessman \small{(2.5\%)} \\
    \textbf{Random Example} & \\
    \hspace{.1cm} M.S.I. Airport is a airport. &  \\
    \hspace{.1cm} Rodmarton is a \mask . & town \small{(16.9\%)} \\ & \textbf{village} \small{(14.7\%)} \\
    \textbf{Close Example} & \\
    \hspace{.1cm} Nantmor is a village. &\\
    \hspace{.1cm} Rodmarton is a \mask . & \textbf{village} \small{(75.5\%)} \\ & hamlet \small{(16.0\%)} \\
    \textbf{Arrow Operator} & \\
    \hspace{.1cm} Totopara $\rightarrow$ village & \\
    \hspace{.1cm} The argument $\rightarrow$ album &\\
    \hspace{.1cm} Tisza $\rightarrow$ river &\\
    \hspace{.1cm} Rodmarton $\rightarrow$ \mask & \textbf{village} \small{(21.4\%)}\\ & town \small{(8.7\%)} \\
    \bottomrule
    \end{tabular}

\caption{Example queries with predictions (from BERT-large) for the different querying methods. The correct answer is marked in bold.}
\label{tab:examples}
\end{table}

\footnotetext{A village in South West England.}

Since we only adjust the context fed to the model, we do not incur the cost of additional forward passes. When paraphrasing, on the other hand, each individual template requires another query to the model. Moreover, our approach does \emph{not} require any learning, i.e., backward passes, and hence is very different from the classic fine-tuning approach and pattern-exploiting training \citep{schick2020exploiting, smalllmfewshot}. 

In Table~\ref{tab:examples}, we compare different approaches of querying by example. The left column shows the input to the model, i.e., the query. The right column shows BERT-large's top-2 prediction, with its corresponding probabilities\footnote{The probabilities are obtained by applying a softmax on the logit output over the token vocabulary.}. The first row of the table shows that completing the \textit{is-a} relation for the village Rodmarton is tricky for the model. Its top predictions are not even close to the correct answer suggesting that BERT either does not know about this particular village or that the information need is not well enough specified. Interestingly, when prepending the query with another \emph{random} example of the same relation (2nd row), the model's top predictions are \textit{town} and the ground-truth \textit{village}. This proves that BERT knows the type of instance Rodmarton is; only the extraction method (the cloze-style template) was not expressive enough. \\

\paragraph{Close Examples} When humans use examples, they typically do not use a completely random subject but use one that is, by some measure, close to the subject at hand. In our introductory example, we used Ronaldo to exemplify an information need about Neuer. It would have been unnatural to use a musician here, even when describing a formally correct \textit{plays-for} relation with them. We extend our approach by only using examples for which the subject is close in latent space to the subject querying for. We use the cosine similarity between the subject encodings using BERT-base. More formally, we encode a subject $s$ using
\begin{align}
    f_\theta (s) = B_\theta(\text{[CLS]} + s + \text{[SEP]})^{\text{CLS}},
\end{align}
with $B(x)^{\text{CLS}}$ being the BERT encoding of the CLS-token for the input $x$, and $\theta$ being the BERT model's parameters. We then obtain the top-$k$ most similar subjects to $s$ in the dataset $\mathcal{D}$ through maximizing the cosine similarity, i.e., 

\begin{align}
    \mathcal{D}' = \arg \max_{\mathcal{D}' \subset \mathcal{D} \setminus \{s\}, |\mathcal{D'}| =k} \sum_{s' \in \mathcal{D'}} \frac{f_\theta(s)^\top f_\theta(s')}{\|f_\theta (s)\|\|f_\theta (s')\|}
\end{align}

From the top-$k$ subset of most similar subjects $\mathcal{D}'$, we randomly sample to obtain our priming examples. Table~\ref{tab:examples} (3rd row) shows the chosen close example to Rodmarton, which is Nantmor, another small village in the UK. Provided with this particular example, BERT-large predicts the ground-truth label \textit{village} with more than 75\% probability. 

\paragraph{Arrow Operator} \citet{gpt3} propose to use LMs as in-context learners. They suggest providing "training" examples in the model's context using the arrow operator, i.e., to express an $\langle s, r, o\rangle$ triplet they provide the model with $s \Rightarrow o$.
We can apply this concept to the LAMA data by using the same template $t_r = $" [$s$] $\Rightarrow$ [$o$]" $\forall r$. In Table \ref{tab:examples} (last row), we see that by providing a few examples of the \textit{is-a} relation, BERT-large can rank the ground-truth highest even though the relationship is never explicitly described in natural language. However, not using a natural language template makes the model less confident in its prediction, as can be seen by the lower probability mass it puts on the target.

\section{Results}
We focus the reporting of the results on the mean precision at k (P@k) metric. In line with previous work \citep{lama, petroni2020context, whatlmknow}\footnote{The P@1 score corresponds to \citet{whatlmknow}'s micro-averaged accuracy}, we compute the results per relation and then average across all relations of the dataset. More formally, for the dataset $\mathcal{D} = \{\mathcal{R}_1, \dots, \mathcal{R}_n\}$ that consists of $n$ relations where each relation has multiple datapoints $\langle x, y \rangle$, we compute the P@k score as: 

\begin{align}
    P@k = \frac{1}{|\mathcal{D}|} \sum_{\mathcal{R}_i \in \mathcal{D}} \frac{1}{|\mathcal{R}_i|} \sum_{\langle x, y\rangle \in \mathcal{R}_i} \mathbbm{1}_{\mathcal{V}'_x}(y),
\end{align}
where $\mathbbm{1}$ denotes the indicator function that is $1$ if the ground truth $y$ is in the top-k prediction set $\mathcal{V}'$ for the input $x$ and $0$ otherwise.\\

\begin{table*}[t]
\centering
\resizebox{\textwidth}{!}{

    \begin{tabular}{llcccccccccccc}
    \toprule
    \multirow{2}{*}{Corpus} & \multirow{2}{*}{Relation} & \multicolumn{5}{|c}{Baselines} & \multicolumn{7}{|c}{LM}  \\
             & & Bb & Bl & Al & Bb$_{{opt}}$ & Bl$_{{opt}}$ & Bb$^3$ & Bb$^{10}$ & Bb$^{10}_{ce}$ & Bl$^{3}$ & Bl$^{10}$ & Bl$^{10}_{ce}$ & Al$^{10}_{ce}$  \\
    \midrule
    \multirow{4}{*}{Google-RE}  & birth-place & \cellcolor[rgb]{0.5651364859669358, 0.8175009611687812, 0.5119723183391004} 14.9 & \cellcolor[rgb]{0.4672049211841599, 0.7748096885813149, 0.4727873894655901} \textbf{16.1} & \cellcolor[rgb]{1.0, 1.0, 0.8980392156862745} 6.3 & - & - & \cellcolor[rgb]{0.8869665513264129, 0.9555709342560553, 0.6656055363321799} 10.5 $_{\pm 0.4}$ & \cellcolor[rgb]{0.7034678969627066, 0.8774778931180316, 0.5688119953863898} 13.2 $_{\pm 0.3}$ & \cellcolor[rgb]{0.8171472510572857, 0.9265667051134179, 0.6230680507497116} 11.7 $_{\pm 0.3}$ & \cellcolor[rgb]{0.9644752018454441, 0.9865743944636679, 0.7224452133794694} 8.9 $_{\pm 0.5}$ & \cellcolor[rgb]{0.8279738562091503, 0.9312418300653594, 0.628235294117647} 11.5 $_{\pm 0.3}$ & \cellcolor[rgb]{0.8611303344867358, 0.9452364475201845, 0.6466589773164167} 11.0 $_{\pm 0.3}$ & \cellcolor[rgb]{0.9921261053440984, 0.9970472895040369, 0.8547327950788158} 7.0 $_{\pm 0.3}$ \\
    &birth-date & \cellcolor[rgb]{0.6980545943867744, 0.8751403306420608, 0.5662283737024222} \textbf{1.6} & \cellcolor[rgb]{0.8496270665128797, 0.9405920799692425, 0.6385697808535178} 1.5 & \cellcolor[rgb]{0.8009073433294887, 0.9195540176855056, 0.6153171856978085} 1.5 & - & - & \cellcolor[rgb]{0.9931103421760861, 0.9974163783160324, 0.8601460976547481} 1.1 $_{\pm 0.3}$ & \cellcolor[rgb]{1.0, 1.0, 0.8980392156862745} 1.1 $_{\pm 0.2}$ & \cellcolor[rgb]{0.9753940792003075, 0.9907727797001153, 0.7627066512879661} 1.2 $_{\pm 0.1}$ & \cellcolor[rgb]{0.9238754325259515, 0.9703344867358709, 0.6926720492118416} 1.4 $_{\pm 0.3}$ & \cellcolor[rgb]{0.8906574394463668, 0.957047289504037, 0.6683121876201461} 1.4 $_{\pm 0.2}$ & \cellcolor[rgb]{0.8009073433294887, 0.9195540176855056, 0.6153171856978085} 1.5 $_{\pm 0.1}$ & \cellcolor[rgb]{0.883275663206459, 0.9540945790080738, 0.6628988850442137} 1.4 $_{\pm 0.3}$ \\
    &death-place & \cellcolor[rgb]{0.39277201076509033, 0.7382698961937716, 0.4348942714340638} 13.1 & \cellcolor[rgb]{0.31833910034602075, 0.7017301038062284, 0.39700115340253744} \textbf{14.0} & \cellcolor[rgb]{1.0, 1.0, 0.8980392156862745} 2.0 & - & - & \cellcolor[rgb]{0.708881199538639, 0.8798154555940023, 0.5713956170703576} 9.2 $_{\pm 0.5}$ & \cellcolor[rgb]{0.4999307958477509, 0.789204152249135, 0.48613610149942327} 11.8 $_{\pm 0.7}$ & \cellcolor[rgb]{0.6173010380622838, 0.8401384083044983, 0.532641291810842} 10.4 $_{\pm 1.0}$ & \cellcolor[rgb]{0.8442137639369474, 0.9382545174932718, 0.6359861591695501} 7.2 $_{\pm 0.7}$ & \cellcolor[rgb]{0.7142945021145714, 0.8821530180699731, 0.5739792387543252} 9.1 $_{\pm 0.5}$ & \cellcolor[rgb]{0.75760092272203, 0.9008535178777394, 0.5946482122260669} 8.5 $_{\pm 1.1}$ & \cellcolor[rgb]{0.9460207612456747, 0.9791926182237601, 0.7089119569396386} 5.0 $_{\pm 0.6}$ \\
    \cmidrule{2-14}
    & Total & \cellcolor[rgb]{0.5586159169550174, 0.8146712802768167, 0.5093886966551326} 9.9 & \cellcolor[rgb]{0.4868896578239139, 0.7835447904652058, 0.4809688581314879} 10.5 & \cellcolor[rgb]{1.0, 1.0, 0.8980392156862745} 3.3 & \cellcolor[rgb]{0.4999307958477509, 0.789204152249135, 0.48613610149942327} 10.4 & \cellcolor[rgb]{0.39953863898500575, 0.7415916955017301, 0.43833910034602075} \textbf{11.3} & \cellcolor[rgb]{0.8537485582468282, 0.9422837370242214, 0.6412456747404844} 6.9 $_{\pm 0.1}$ & \cellcolor[rgb]{0.6872279892349097, 0.8704652056901192, 0.5610611303344868} 8.7 $_{\pm 0.2}$ & \cellcolor[rgb]{0.7792541330257593, 0.9102037677816225, 0.6049826989619377} 7.8 $_{\pm 0.4}$ & \cellcolor[rgb]{0.9275663206459054, 0.9718108419838524, 0.6953787004998078} 5.8 $_{\pm 0.4}$ & \cellcolor[rgb]{0.8171472510572857, 0.9265667051134179, 0.6230680507497116} 7.4 $_{\pm 0.1}$ & \cellcolor[rgb]{0.8496270665128797, 0.9405920799692425, 0.6385697808535178} 7.0 $_{\pm 0.4}$ & \cellcolor[rgb]{0.9803152633602461, 0.9926182237600923, 0.7897731641676278} 4.5 $_{\pm 0.3}$ \\
    \midrule
    \multirow{4}{*}{T-REx}  & $1$-$1$ & \cellcolor[rgb]{0.9570934256055363, 0.9836216839677048, 0.7170319108035371} 68.0 & \cellcolor[rgb]{0.8537485582468282, 0.9422837370242214, 0.6412456747404844} \textbf{74.5} & \cellcolor[rgb]{0.9054209919261822, 0.9629527104959631, 0.6791387927720107} 71.2 & - & - & \cellcolor[rgb]{1.0, 1.0, 0.8980392156862745} 59.7 $_{\pm 0.6}$ & \cellcolor[rgb]{0.9911418685121107, 0.9966782006920415, 0.8493194925028835} 62.0 $_{\pm 0.6}$ & \cellcolor[rgb]{0.9881891580161476, 0.9955709342560554, 0.8330795847750865} 62.6 $_{\pm 0.8}$ & \cellcolor[rgb]{0.9724413687043445, 0.9896655132641292, 0.7464667435601692} 66.4 $_{\pm 0.9}$ & \cellcolor[rgb]{0.9644752018454441, 0.9865743944636679, 0.7224452133794694} 67.6 $_{\pm 0.6}$ & \cellcolor[rgb]{0.9460207612456747, 0.9791926182237601, 0.7089119569396386} 68.7 $_{\pm 0.7}$ & \cellcolor[rgb]{0.938638985005767, 0.976239907727797, 0.7034986543637063} 69.0 $_{\pm 0.7}$ \\
    &$N$-$1$ &  \cellcolor[rgb]{0.9534025374855825, 0.9821453287197233, 0.7143252595155709} 32.4 & \cellcolor[rgb]{0.9201845444059977, 0.9688581314878894, 0.6899653979238755} 34.2 & \cellcolor[rgb]{1.0, 1.0, 0.8980392156862745} 24.9 & - & - & \cellcolor[rgb]{0.9534025374855825, 0.9821453287197233, 0.7143252595155709} 32.3 $_{\pm 0.1}$ & \cellcolor[rgb]{0.8496270665128797, 0.9405920799692425, 0.6385697808535178} 37.9 $_{\pm 0.2}$ & \cellcolor[rgb]{0.7521876201460977, 0.8985159554017685, 0.5920645905420993} 41.7 $_{\pm 0.4}$ & \cellcolor[rgb]{0.8279738562091503, 0.9312418300653594, 0.628235294117647} 38.8 $_{\pm 0.2}$ & \cellcolor[rgb]{0.6629450211457133, 0.8599461745482507, 0.550726643598616} 44.8 $_{\pm 0.2}$ & \cellcolor[rgb]{0.5651364859669358, 0.8175009611687812, 0.5119723183391004} \textbf{47.9} $_{\pm 0.2}$ & \cellcolor[rgb]{0.6564244521337947, 0.8571164936562861, 0.5481430219146483} 45.0 $_{\pm 0.2}$ \\
    & $N$-$M$ &  \cellcolor[rgb]{0.9201845444059977, 0.9688581314878894, 0.6899653979238755} 24.7 & \cellcolor[rgb]{0.9201845444059977, 0.9688581314878894, 0.6899653979238755} 24.8 & \cellcolor[rgb]{1.0, 1.0, 0.8980392156862745} 17.2 & - & - & \cellcolor[rgb]{0.8496270665128797, 0.9405920799692425, 0.6385697808535178} 27.9 $_{\pm 0.4}$ & \cellcolor[rgb]{0.741361014994233, 0.893840830449827, 0.5868973471741638} 31.3 $_{\pm 0.2}$ & \cellcolor[rgb]{0.6107804690503653, 0.8373087274125337, 0.5300576701268743} 34.8 $_{\pm 0.1}$ & \cellcolor[rgb]{0.7359477124183007, 0.8915032679738563, 0.5843137254901961} 31.4 $_{\pm 0.4}$ & \cellcolor[rgb]{0.6042599000384468, 0.834479046520569, 0.5274740484429066} 35.0 $_{\pm 0.1}$ & \cellcolor[rgb]{0.5194925028835063, 0.7976931949250289, 0.4938869665513264} \textbf{37.2} $_{\pm 0.3}$ & \cellcolor[rgb]{0.6629450211457133, 0.8599461745482507, 0.550726643598616} 33.5 $_{\pm 0.2}$ \\
    \cmidrule{2-14} 
    & Total & \cellcolor[rgb]{0.9497116493656286, 0.9806689734717416, 0.7116186082276048} 31.1 & \cellcolor[rgb]{0.9238754325259515, 0.9703344867358709, 0.6926720492118416} 32.5 & \cellcolor[rgb]{1.0, 1.0, 0.8980392156862745} 24.2 & \cellcolor[rgb]{0.7467743175701653, 0.8961783929257978, 0.5894809688581315} 39.6 & \cellcolor[rgb]{0.6107804690503653, 0.8373087274125337, 0.5300576701268743} 43.9 & \cellcolor[rgb]{0.9349480968858132, 0.9747635524798155, 0.7007920030757401} 31.9 $_{\pm 0.2}$ & \cellcolor[rgb]{0.838800461361015, 0.9359169550173011, 0.6334025374855825} 36.5 $_{\pm 0.2}$ & \cellcolor[rgb]{0.7359477124183007, 0.8915032679738563, 0.5843137254901961} 40.0 $_{\pm 0.2}$ & \cellcolor[rgb]{0.8171472510572857, 0.9265667051134179, 0.6230680507497116} 37.3 $_{\pm 0.2}$ & \cellcolor[rgb]{0.6759861591695502, 0.86560553633218, 0.5558938869665514} 42.1 $_{\pm 0.2}$ & \cellcolor[rgb]{0.5781776239907728, 0.8231603229527105, 0.5171395617070358} \textbf{44.8} $_{\pm 0.1}$ & \cellcolor[rgb]{0.6872279892349097, 0.8704652056901192, 0.5610611303344868} 41.7 $_{\pm 0.1}$ \\
    \midrule 
    ConceptNet & Total & \cellcolor[rgb]{0.9931103421760861, 0.9974163783160324, 0.8601460976547481} 15.9 & \cellcolor[rgb]{0.938638985005767, 0.976239907727797, 0.7034986543637063} 19.5 & \cellcolor[rgb]{0.8758938869665513, 0.9511418685121107, 0.6574855824682814} 21.2 & - & - & \cellcolor[rgb]{1.0, 1.0, 0.8980392156862745} 15.2 $_{\pm 0.2}$ & \cellcolor[rgb]{0.9911418685121107, 0.9966782006920415, 0.8493194925028835} 16.2 $_{\pm 0.2}$ & \cellcolor[rgb]{0.9822837370242214, 0.9933564013840831, 0.8005997693194925} 17.1 $_{\pm 0.2}$ & \cellcolor[rgb]{0.9349480968858132, 0.9747635524798155, 0.7007920030757401} 19.6 $_{\pm 0.3}$ & \cellcolor[rgb]{0.8795847750865051, 0.9526182237600923, 0.6601922337562476} 21.2 $_{\pm 0.2}$ & \cellcolor[rgb]{0.8496270665128797, 0.9405920799692425, 0.6385697808535178} 22.0 $_{\pm 0.3}$ & \cellcolor[rgb]{0.6107804690503653, 0.8373087274125337, 0.5300576701268743} \textbf{26.5} $_{\pm 0.2}$ \\
    \bottomrule
    \end{tabular}

}
\caption{Mean precision at one (P@1) in percent across the different corpora of the LAMA probe. The baseline models shown are BERT-base (Bb), BERT-large (Bl), Albert-xxlarge-v2 (Al), and the best versions of BERT-large and BERT-base by \citet{whatlmknow} that are optimized across multiple paraphrases\footnotemark (Bb$_{opt}$ and Bl$_{opt}$). The LM section on the right shows the results for different querying by example approaches. Here, the superscript denotes the number of examples used and the subscript \textit{ce} denotes that only close examples have been used. Since the choice of examples alters the predictions of the model and thus introduces randomness, we provide the standard deviation measured over 10 evaluations.}
\label{tab:results}
\end{table*}


Table~\ref{tab:results} shows the P@1 scores of different models and querying approaches across the LAMA probe's corpora. While for the Google-RE data, providing additional examples shows to be detrimental, we see massive prediction performance gains for T-REx and ConceptNet. Most notably, the P@1 score of BERT-large on T-REx increases by 37.8\% to 44.8\% when providing 10 close examples. Similarly, the lower bound on Albert's performance for T-REx (ConceptNet) can be improved by up to 72.3\% (25.0\%) with 10 close examples. 

\paragraph{Google-RE} For the Google-RE subset of the data, querying by example hurts the predictive capabilities of LMs. In the following, we provide an intuition of why we think this is the case. Looking at the baseline numbers of the individual relations for this data, we see that the performance is largely driven by predicting a person's birth and death place; the birth-date relation doesn't play a significant role because BERT is incapable of accurately predicting numbers (i.e., dates) \citep{lin2020birds, wallace2019nlp}. The birth and death place of a person BERT-large predicts correctly 16.1\% and 14.0\% of the time, respectively; significantly lower than the 32.5\% P@1 score among the relations of the T-REx data. Recent work describes that BERT has a bias to predict that a person with, e.g., an Italian sounding name is Italian \citep{bertology, poerner2020ebert}. We suspect that this bias helps BERT predict birth and death places without knowing the actual person, and therefore it is not an adequate test of probing an LMs factual knowledge. As a consequence, the predictions it makes are more prone to errors when influenced by previous examples.

\begin{figure}
    \centering
    \includegraphics[width=0.4\textwidth]{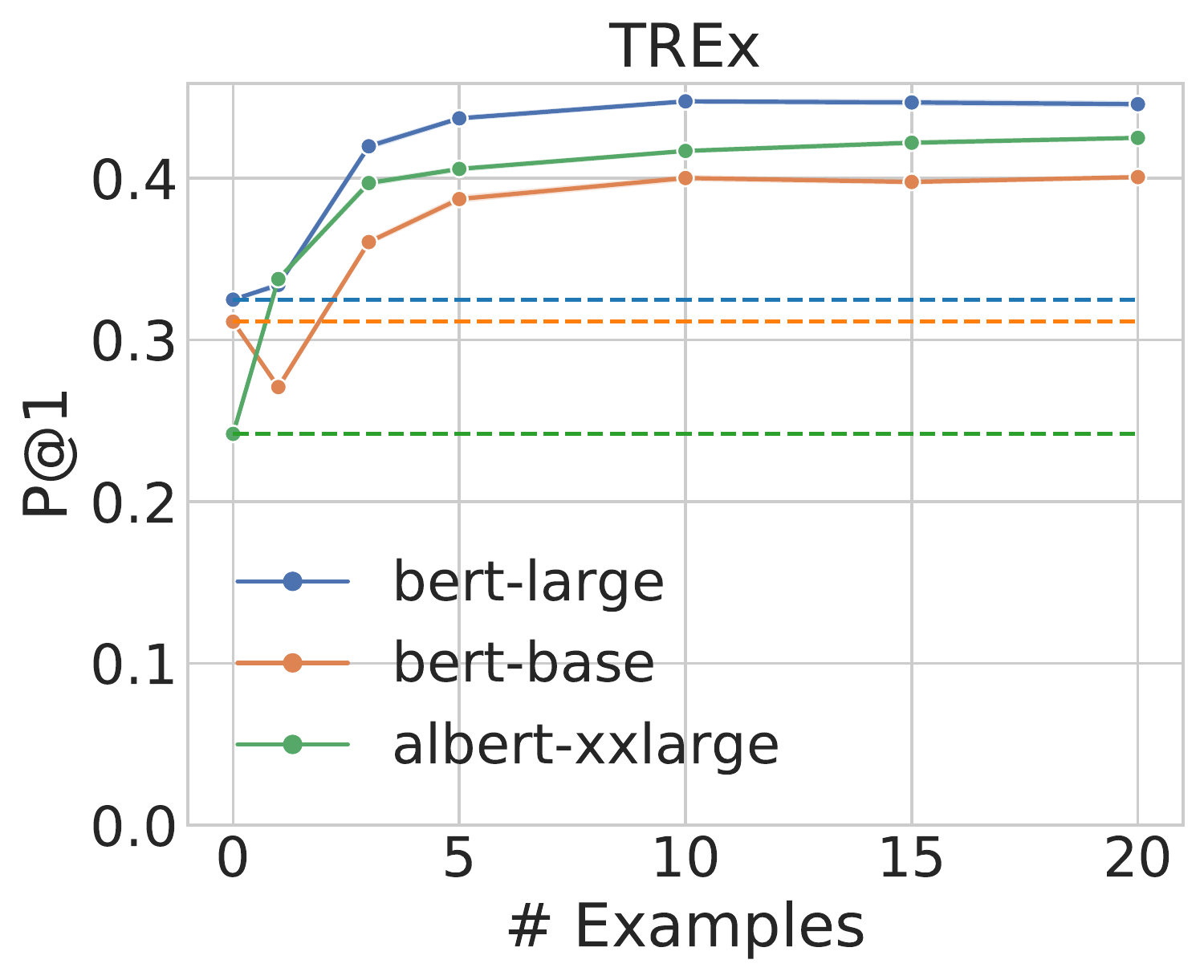}
    \caption{P@1 score for TREx over the number of examples provided. The dashed line shows the baseline value for when no additional example is given.}
    \label{fig:p@1}
\end{figure}

\paragraph{T-REx} Figure~\ref{fig:p@1} depicts the mean precision at 1 on the T-REx corpus for a varying number of examples provided. It shows that even a few additional examples can significantly improve the performance of the LMs. However, there is a saturation of usefulness for more examples that seems to be reached at around 10 examples already. Interestingly, with 10 examples, BERT-large even slightly improves upon the optimized paraphrase baseline from \citet{whatlmknow}, while only requiring a single forward pass.\\
Table~\ref{tab:gain} shows the improvement in P@1 score for the individual relations that most (and least) benefit from additional examples for BERT-large. The relations for which demonstrations improve the performance the most typically have one thing in common: they are ambiguous. Prototypical ambiguous relations like \textit{located-in} or \textit{is-a} are among the top benefiting relations. One rather untypical improvement candidate is the top-scoring one of \textit{religion-affiliation}. Suspiciously, this is also the most improved relation by the paraphrasing of \citet{whatlmknow}. A closer look at the examples reveals the cause: the target object labels for the religions are provided as nouns (e.g., Christianity, Islam), while the template (\textit{[s] is affiliated with the [o] religion}) indicates to use the religion as an adjective (e.g., Christian, Islamic). Hence, both paraphrasing the sentence such that it is clear to use a noun or providing example sentences that complete the template with nouns alleviate this problem. The relations that benefit the least from demonstrations are unambiguous, like \textit{capital-of} or \textit{developed-by}.

\begin{table}[h]
\centering
\resizebox{0.5\textwidth}{!}{   
 \begin{tabular}{llccc}
    \toprule
    \multirow{2}{*}{\textbf{ID}} & \multirow{2}{*}{\textbf{Template}} & \multicolumn{3}{c}{\textbf{$\Delta$ P@1}} \\
    & & n=1 & n=3 & n=5 \\
    \midrule
P140 & [s] is affiliated with the [o] religion . & 51.0 & 67.4 & 70.0 \\
P30 & [s] is located in [o] . & 47.8 & 55.3 & 55.8 \\
P136 & [s] plays [o] music . & 12.8 & 44.0 & 54.5 \\
P31 & [s] is a [o] . & 8.2 & 20.3 & 24.4 \\
\dots & & & &\\
P178 & [s] is developed by [o] . & -8.3 & -4.2 & -6.8 \\
P1376 & [s] is the capital of [o] . & -16.3 & -8.2 & -8.6 \\
\bottomrule
\end{tabular}
}
\caption{List of relations of T-REx that benefit the most (least) by additional examples. The right column provides the improvement in precision at 1 score when \{1, 3, 5\} examples are provided for BERT-large.}
\label{tab:gain}
\end{table}

\paragraph{ConceptNet} While T-REx probes for factual knowledge, the ConceptNet corpus is concerned with commonsense relations. The improvements of querying by example are significant with 12\%, 7.5\%, and 25\% relative improvement for BERT-base, BERT-large, and Albert-xxlarge. \\

More detailed plots for all the corpora and several metrics are provided in Appendix~\ref{sec:omitted_figures}.

\subsection{The Change of Embedding}
\begin{figure*}
    \centering
    \hspace*{-0.05\textwidth}
    \includegraphics[width=.9\textwidth]{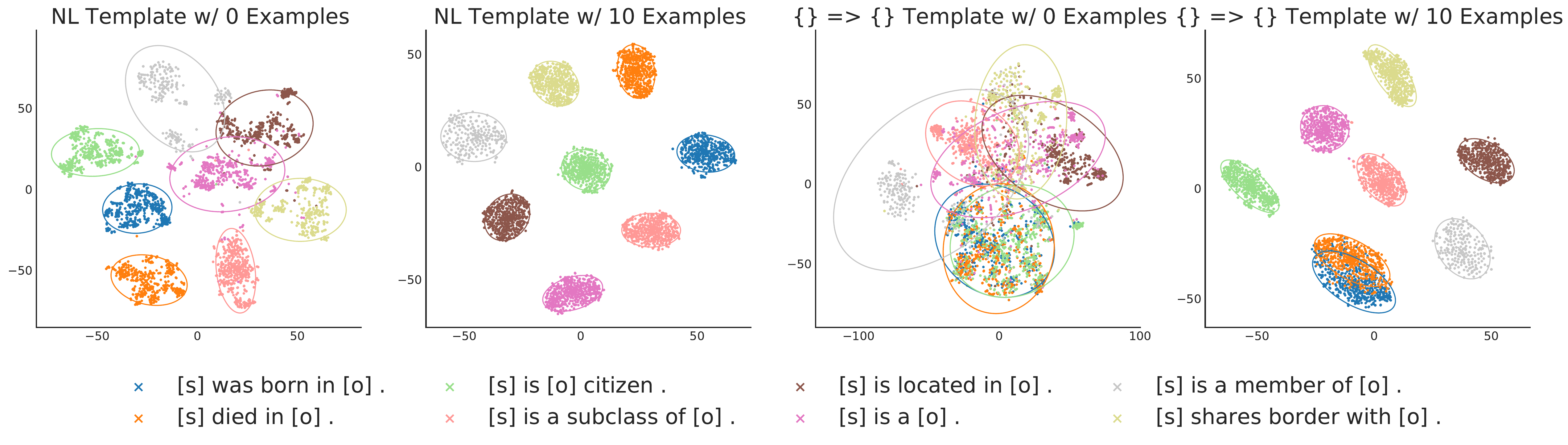}
    \caption{BERT-large's [CLS]-token embedding of a subset of T-REx queries visualized in two dimensions using t-SNE \citep{JMLR:v9:vandermaaten08a}. Each point is a single query and the color represents the corresponding relation class. The ellipses depict the 2-std confidence intervals. The individual images show the clustering for both the natural language and the ([s]; [o]) template with either no examples or ten examples provided.}
    \label{fig:embeddings}
\end{figure*}

To further investigate the disambiguation effect of additional examples, we take a look at the latent space. In particular, we're interested in how the clusters of particular relations, formed by the queries' embeddings, change when providing the context with additional examples. Figure~\ref{fig:embeddings} visualizes BERT-large's [CLS]-token embedding for queries from the T-REx corpus, using t-SNE \citep{JMLR:v9:vandermaaten08a}. The individual colors represent the relations of the queries. The first two images depict the clustering when using the natural language template without additional demonstrations (left) and ten demonstrations (right). The fact that the clusters become better separated is visual proof that providing examples disambiguates the information need expressed by the queries. The two plots on the right show the clustering when instead of a natural language template, the subject and object are only separated by the arrow operator "$\Rightarrow$". Here, we see an even more significant change in separability when providing additional demonstrations, as the actual information need is more ambiguous.

\footnotetext{These models involve one query to the model per paraphrase.}

\subsection{TextWorld Commonsense Evaluation}
\begin{figure}
    \centering
    \includegraphics[width=0.4\textwidth]{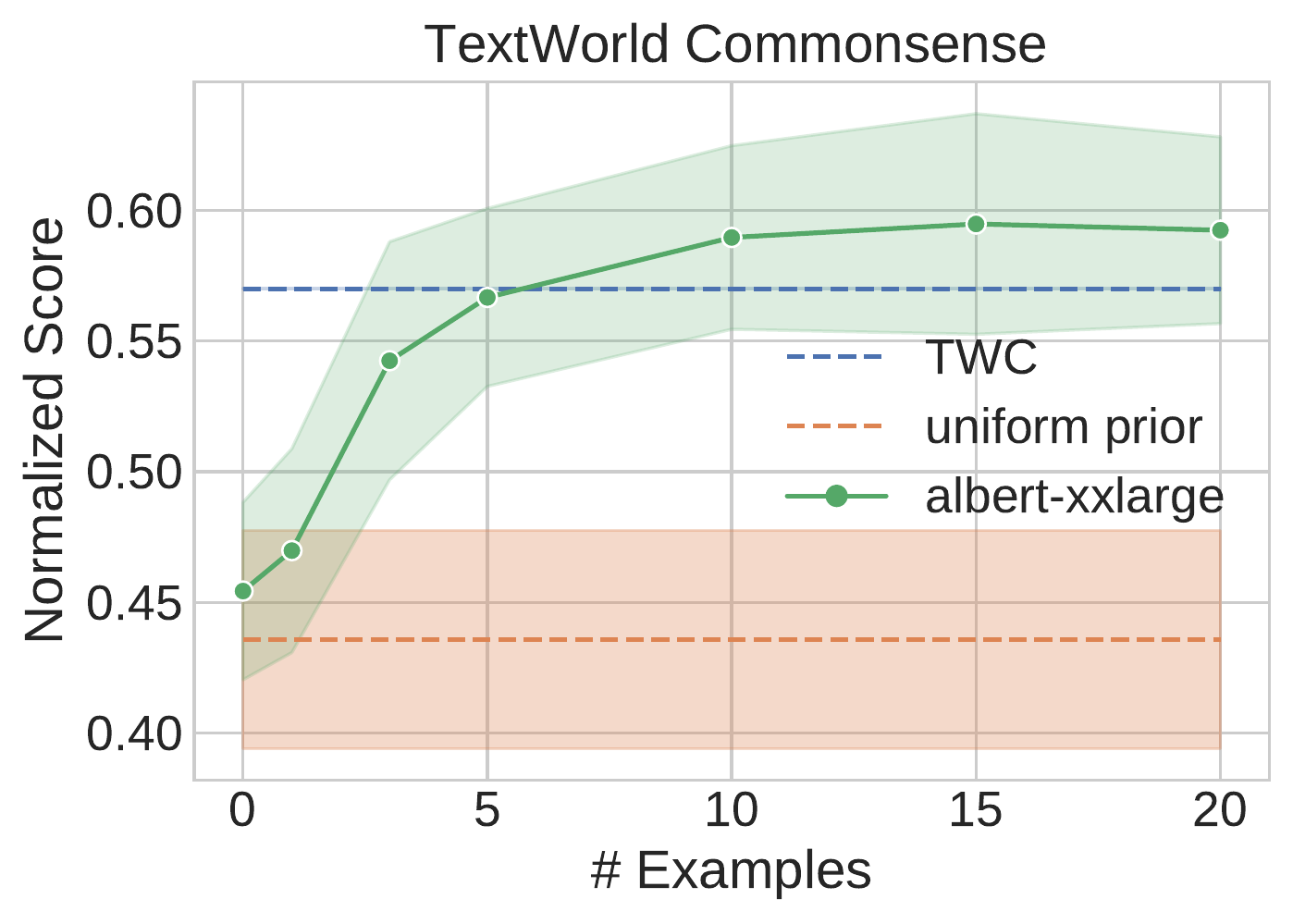}
    \caption{Normalized score for the \textit{hard} games of the \texttt{TWC} environment over the number of examples provided for albert-xxlarge. The dashed baselines are the static agent with a uniform prior and the TWC commonsense agent by \citet{murugesan2020textbased}. The shaded regions depict the standard deviation over 10 runs.}
    \label{fig:twc}
\end{figure}

An emerging field of interest inside the NLP community is text-based games (TBG). An agent is placed inside an interactive text environment in these games and tries to complete specified goals--only using language commands. To succeed, it requires a deep language understanding to decide what are reasonable actions to take in the scene that move it closer to its final goal. These environments are often modeled on real-world scenes to foster the commonsense-learning capabilities of an agent. The TextWorld Commonsense (\texttt{TWC}) game world by \citet{murugesan2020textbased} focus specifically on this aspect. There, the agent is placed in a typical modern-house environment to tidy up the room. This involves moving all the objects in the scene to their \textit{commonsense} location, e.g., the dirty dishes belong in the dishwasher and not in the cupboard. \citet{murugesan2020textbased} approach this problem by equipping the agent with access to a commonsense knowledge base. Replacing a traditional KB with an LM for this task is very intriguing as the LM has relational knowledge stored implicitly and is capable of generalizing to similar objects. To test the feasibility of using LMs as commonsense knowledge source in the \texttt{TWC} environment, we design the following experiment\footnote{Details and the pseudocode are provided in Apendix \ref{app:twc}}: We use a static agent that picks up any misplaced object $o$ at random and puts it to one of the possible locations $l$ in the scene according to a specific prior $p(l|o)$. This prior $p(l|o)$ is computed at the start of an episode for all object-location combinations in the scene, using an LM. We use the arrow operator as described in Table \ref{tab:examples} and vary the number of examples provided. In Figure \ref{fig:twc}, we show the result for albert-xxlarge on the \textit{hard} games of \texttt{TWC}, compared to a simple uniform prior (i.e., $p(l_i|o) = const. \, \forall i$), and \citet{murugesan2020textbased}'s RL agent with access to a commonsense KB. We see the same trend as in the LAMA experiments: providing additional examples of the same relation boosts performance significantly and saturates after 10-15 instances.

\subsection{Word Analogy Evaluation}
To evaluate the usefulness of querying pre-trained language models by examples for linguistic knowledge, we move to the word analogy task---a standard benchmark for non-contextual word embeddings. This evaluation is based on the premise that a good global word embedding defines a latent space in which basic arithmetic operations correspond to linguistic relations \citep{mikolov-etal-2013-linguistic}. With the rise of contextual word embeddings and large pre-trained language models, this evaluation has lost significance. However, we consider approaching this task from the angle of querying linguistic knowledge from an LM instead of performing arithmetics in latent space. By providing examples of the linguistic relation with a regular pattern in the context of the LM, we prime it to apply the relation to the final word with its masked out correspondence. \\
We consider the \textit{Bigger Analogy Test Set (BATS)} \citep{GladkovaDrozd2016} for our experiments. BATS consists of 40 different relations covering inflectional and derivational morphology, as well as lexicographic and encyclopedic semantics. Each relation consists of 50 unique word pairs. However, since most pre-trained LMs, including BERT and Albert, use subword-level tokens for their vocabulary, not all examples can be solved. In particular, 76.1\% and 76.2\% of the targets are contained in BERT's and Albert's vocabulary, respectively---upper bounding their P@1 performance. \\
\begin{figure}
    \centering
    \includegraphics[width=0.4\textwidth]{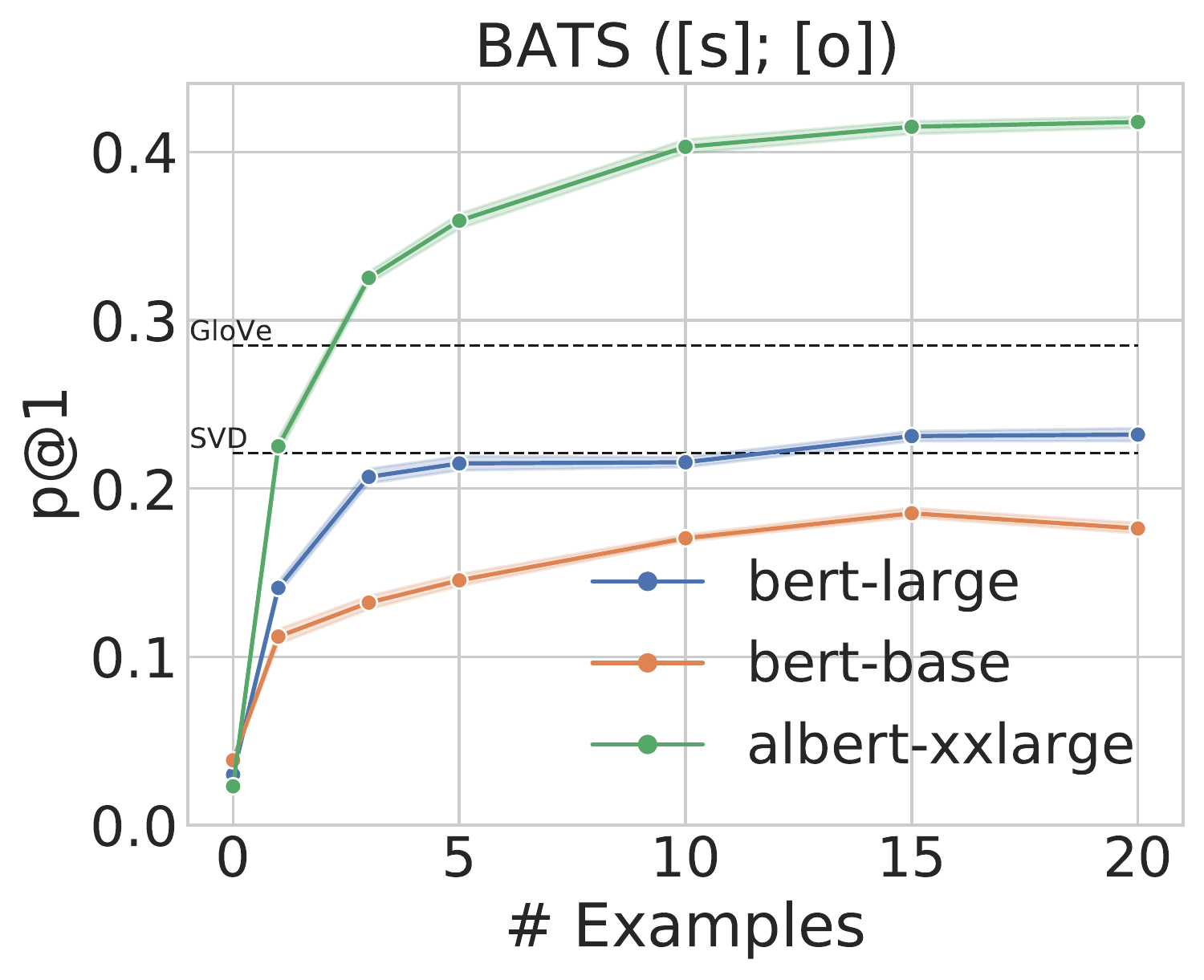}
    \caption{P@1 score on BATS over the number of examples provided. The performance of the GloVe and SVD benchmark models by \citet{GladkovaDrozd2016} is shown with the black, dashed lines.}
    \label{fig:bats_results}
\end{figure}
Figure \ref{fig:bats_results} depicts the P@1 score\footnote{The P@1 score corresponds to \citet{GladkovaDrozd2016}'s reported accuracy score.} for the individual LMs on BATS. Noticeably, also on this task, the LMs benefit from additional examples up to a certain threshold for which the usefulness stagnates. Both BERT models do not beat \citet{GladkovaDrozd2016}'s GloVe \citep{pennington-etal-2014-glove} benchmark. This is in part because not all targets are present in the token vocabulary. Considering only the \textit{solvable} word pairs, BERT-large achieves a P@1 score of 30.6\% with 15 examples---beating the GloVe baseline achieving 28.5\%.
Interestingly, Albert-xxlarge outperforms all other models, including the baselines, by a large margin. Figure~\ref{fig:bats_category} in Appendix~\ref{sec:omitted_figures} breaks down the LM's performance across the different relations of BATS and compares it against the GloVe baseline. Albert beats GloVe on almost all relations where its vocabulary does not limit it; the most significant improvements are in the derivational morphology and lexicographic semantics categories. It is outperformed by GloVe only on two relations: \textit{country:capital} and \textit{UK city:county}. Especially the former \textit{country:capital} category is very prominent and constituted 56.7\% of all semantic questions of the original Google test set \citep{mikolov2013efficient}---potentially influencing the design and tuning of non-contextual word embeddings.

\section{Discussion}
Augmenting the context of LMs with demonstrations is a very successful strategy to disambiguate the query. Notably, it is as successful, on TRE-x, as using an ensemble of multiple paraphrases. The benefit of additional examples decreases when the information need is clear to the model; this is the case for unambiguous prompts or when enough (around 10) demonstrations are provided. Even in the extreme case of ambiguity, for example, when the arrow operator (\textit{[s]} $=>$ \textit{[o]}) is used to indicate a relation, providing only a handful of examples clarifies the relation sufficiently in many cases.
We showed that the usefulness of providing additional demonstrations quickly vanishes. Hence, when having access to more labeled data and the option to re-train the model, a fine-tuning strategy is still better suited to maximize the performance on a given task. Moreover, casting NLP problems as language modeling tasks only works as long as the target is a single-token word of the LM's vocabulary. While technically large generation-based LMs as GPT \citep{gpt3, gpt2} or T5 \citep{T5} can generate longer sequences, it is not clear how to compare solutions of varying length.

\section{Conclusion}
In this work, we explored the effect of providing examples to probing LMs relational knowledge. We showed that already a few demonstrations---supplied in the context of the LM---disambiguate the query to the same extent as using an optimized ensemble of multiple paraphrases. We base our findings on experimental results of the LAMA probe, the BATS word analogy test, and a TBG commonsense evaluation. On the T-REx corpus' factual relations, providing 10 demonstrations improves BERT's P@1 performance by 37.8\%. Similarly, on ConceptNet's commonsense relations, Albert's performance improves by 25\% with access to 10 examples. 
We conclude that providing demonstrations is a simple yet effective strategy to clarify ambiguous prompts to a language model.

\clearpage

\bibliography{ref}
\bibliographystyle{acl_natbib}

\clearpage

\appendix
\onecolumn
\section{Appendices}
\label{sec:appendix}

\subsection{Implementation Details}\label{sec:implementation_details}

The source code to reproduce all the experiments is available at \url{https://github.com/leox1v/lmkb_public}.
All individual runs reported in the paper can be carried out on a single GPU (TESLA P100 16GB), though speedups can be realized when using multiple GPUs in parallel. The wall-clock runtime for the corpora of the LAMA probe is shown in Table~\ref{tab:runtime}. \\
All models used in this work are accessed from the Huggingface's list of pre-trained models for PyTorch \citep{DBLP:journals/corr/abs-1910-03771}. Further details about these models are provided on the following webpage: \url{https://huggingface.co/transformers/pretrained_models.html}.

\begin{table}[h]
\centering
    \begin{tabular}{llccc}
    \toprule
    Corpus & Model & \# Parameters & Avg. Input Length & Runtime [s] \\ 
    \midrule
    \multirow{9}{*}{Google-RE} & bert-base-cased & \multirow{3}{*}{109M} & 5.5 & 12.8 \\
    & bert-base-cased$^{10}$ &  &  60.3 & 36.1 \\
    & bert-base-cased$^{10}_{\text{ce}}$ &  & 60.1 & 39.6 \\
    \cmidrule{2-5}
    & bert-large-cased & \multirow{3}{*}{335M} & 5.5 & 20.5 \\
    & bert-large-cased$^{10}$ & & 60.3 & 85.5 \\
    & bert-large-cased$^{10}_{\text{ce}}$ & & 60.1 & 99.7 \\
    \cmidrule{2-5}
    & albert-xxlarge-v2 & \multirow{3}{*}{223M} & 5.5 & 85.4 \\
    & albert-xxlarge-v2$^{10}$ & & 60.3 & 466.0 \\
    & albert-xxlarge-v2$^{10}_{\text{ce}}$ & & 60.1 & 544.9 \\
    \hline
    \multirow{9}{*}{T-REx} & bert-base-cased & \multirow{3}{*}{109M} & 7.6 & 72.6 \\
    & bert-base-cased$^{10}$ &  & 83.2 & 239.0 \\
    & bert-base-cased$^{10}_{\text{ce}}$ &  & 82.7 & 234.1 \\
    \cmidrule{2-5}
    & bert-large-cased & \multirow{3}{*}{335M} & 7.6 & 119.3 \\
    & bert-large-cased$^{10}$ & & 83.2 & 747.5 \\
    & bert-large-cased$^{10}_{\text{ce}}$ & & 82.7 & 596.5 \\
    \cmidrule{2-5}
    & albert-xxlarge-v2 & \multirow{3}{*}{223M} & 7.6 & 504.1 \\
    & albert-xxlarge-v2$^{10}$ & & 83.2 & 3227.4 \\
    & albert-xxlarge-v2$^{10}_{\text{ce}}$ & & 82.7 & 3340.9 \\
    \hline
    \multirow{9}{*}{ConceptNet} & bert-base-cased & \multirow{3}{*}{109M} & 9.4 & 38.5 \\
    & bert-base-cased$^{10}$ &  & 102.8 & 121.9 \\
    & bert-base-cased$^{10}_{\text{ce}}$ &  & 104.5 & 124.6 \\
    \cmidrule{2-5}
    & bert-large-cased & \multirow{3}{*}{335M} & 9.4 & 80.4 \\
    & bert-large-cased$^{10}$ & & 102.8 & 311.4 \\
    & bert-large-cased$^{10}_{\text{ce}}$ & & 104.5 & 324.3 \\
    \cmidrule{2-5}
    & albert-xxlarge-v2 & \multirow{3}{*}{223M} & 9.4 & 408.0 \\
    & albert-xxlarge-v2$^{10}$ & & 102.8 & 1760.8 \\
    & albert-xxlarge-v2$^{10}_{\text{ce}}$ & & 104.5 & 1853.6 \\
    \bottomrule
    \end{tabular}
\caption{The runtime in seconds to go once through the full data from the LAMA probe on a single TESLA P100 GPU with a batch size of 32. The superscript of the model represents the number of examples used for querying and the subscript of \textit{ce} indicates that close examples are used.}
\label{tab:runtime}
\end{table}

\subsection{The Choice of Template}\label{sec:choice_of_template}
\begin{figure}
    \centering
    \includegraphics[width=0.45\textwidth]{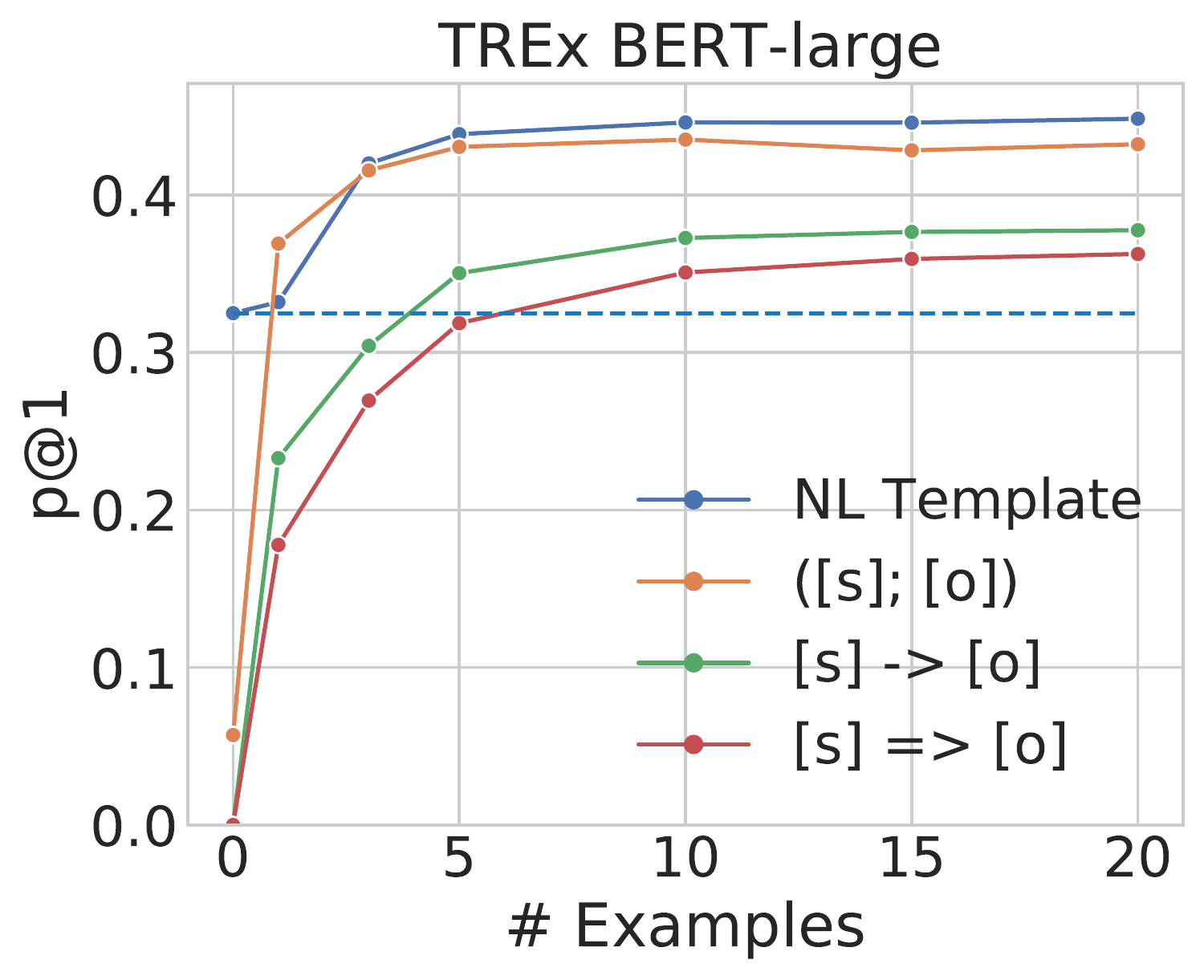}
    \caption{P@1 score for BERT-large on TREx over the number of examples provided. Each line corresponds to one \textit{template} determining how the examples are provided: (i) with the natural language templates from the LAMA probe (NL Template), (ii) separated by a semicolon (([s]; [o])), (iii) separated by a one-lined arrow ([s] -$>$ [o]), or (iv) separated by a double-lined arrow ([s] =$>$ [o]). The dashed line shows the baseline value for when no additional example is given.}
    \label{fig:p@1_arrow}
\end{figure}

When providing examples, we give the model the chance to understand the relationship for which we query without providing additional instructions. This naturally raises the question of whether or not natural language templates are even necessary to query LMs. Most prominently, the in-context learning of \citet{gpt3} shows that large LMs can complete patterns even when not provided in natural language. In particular, they use the "=$>$"-operator to express the relation between input and output. In Figure~\ref{fig:p@1_arrow}, we compare the natural language cloze-style template against three different non-language templates: (i) \textit{[s] =$>$ [o]}, (ii) \textit{[s] -$>$ [o]}, (iii) \textit{([s]; [o])}. Surprisingly, \citet{gpt3}'s "=$>$"-operator performs the worst for BERT-large on T-TREx, while separating the subject and objects by a semicolon works best---almost on par with the performance of the natural language template after providing just a single example. This result underlines BERT's remarkable pattern-matching capabilities and suggests that a natural language description of the relation is not always needed---even when querying relatively small LMs.

\subsection{Details TextWorld Commonsense Evaluation}\label{app:twc}
Text-based games (TBG) are computer games where the sole modality of interaction is text. Classic games like Zork \citep{infocom} used to be played by a large fan base worldwide. Today, they provide interesting challenges for the research field of interactive NLP. With the TextWorld framework by \citet{DBLP:journals/corr/abs-1806-11532}, it is possible to design custom TBGs; allowing to adapt the objects, locations, and goals around the investigated research objectives. TBGs of this framework can vary from treasure hunting \citep{DBLP:journals/corr/abs-1806-11532} to cooking recipes \citep{adhikari2021learning, DBLP:journals/corr/abs-1909-01646}, or--as in the experiment at hand--tidying up a room \citep{murugesan2020textbased}. \citet{murugesan2020textbased} designed the TextWorld Commonsense environment \texttt{TWC} around the task of cleaning up a modern house environment to probe an agent about its commonsense abilities. For example, a successful agent should understand that dirty dishes belong in the dishwasher while clean dishes in the cupboard. \citet{murugesan2020textbased} approach this problem by developing an agent that, through a graph-based network, has access to relevant facts from the ConceptNet \citep{speer2018conceptnet} commonsense knowledge base. Here, the obvious downside of static KBs for commonsense knowledge extraction becomes apparent: it does not generalize to not listed object-location pairs. Hence, slight deviations of typical entities require additional processing to be able to query the KB. A large pre-trained LM seems to be better suited for this task due to its querying flexibility and generalization capabilities. We test these abilities by designing a static agent as described in the following Algorithm \ref{alg:twc}, that has access to a large pre-trained LM. 
\clearpage

\begin{algorithm}
\SetAlgoLined
\DontPrintSemicolon
 \KwIn{\texttt{TWC} game \texttt{G}, pre-trained language model \texttt{LM}}\;
 $o_s \gets \text{objects in the scene}$\;
 $l_s \gets \text{locations in the scene}$\;
 $o \gets \text{large list of all possible objects across all games}$\;
 \;
 
 \SetKwFunction{FMain}{GetPrior}
  \SetKwProg{Fn}{Function}{:}{}
  \Fn{\FMain{$o_s, l_s, o, \texttt{LM}$}}{%
  \tcc{Function to determine a probability distribution over the locations $l_s$ for each object in $o_s$ using the language model $\texttt{LM}$.}
    $p \gets \text{empty array of size } |o_s|\times | l_s|$ \;
    \ForAll{object $o_i \in o_s$}{
        $d \gets \text{Randomly sample demonstrations for objects }\in o \setminus o_s \text{ with locations} \in l_s$ \;\;
        \tcc{Use demonstrations $d$ to build context for \texttt{LM}, e.g.: \\
        milk $\Rightarrow$ fridge\\
        dirty dishes $\Rightarrow$ sink \\
        $o_i \Rightarrow$ [MASK]}
        \;
        $c \gets build\_context(d)$ \;\;
        \tcc{Compute MASK-token probabilities for the locations in $l_s$ using $\texttt{LM}$}
        $p_{o_i} \gets \texttt{LM}(c, l_s)$\;
        p.append($p_{o_i}$)
    }
    \KwRet p
  }
  \;
 $prior \gets \texttt{GetPrior}(o_s, l_s, o, \texttt{LM})$\;\;
 \While{$\texttt{G}$ not finished \&  max steps not exhausted}{
  \eIf{agent holds an object $o_i$}{
    $l_i \gets \text{sample location according to prior}[o_i]$ \\
    \eIf{$l_i$ correct location for $o_i$}{
        remove $o_i$ from $o_s$ \\
    }{
        prior[$o_i$] $\gets$ 0\\
    }
  }{
  $o_i \gets$ random\_choice($o_s$) \\
  }
 }
 \caption{LM-prior Agent}
 \label{alg:twc}
\end{algorithm}

\clearpage

\subsection{Omitted Figures}\label{sec:omitted_figures}
\begin{table}[h!]
\centering
\resizebox{\textwidth}{!}{   
    \begin{tabular}{p{.3cm}c@{}c@{}c@{}}
     & Google-RE & T-REx & ConceptNet \\
    \toprule
    \rotatebox[origin=l]{90}{\hphantom{mmm}Random} & 
    \includegraphics[width=0.25\textwidth]{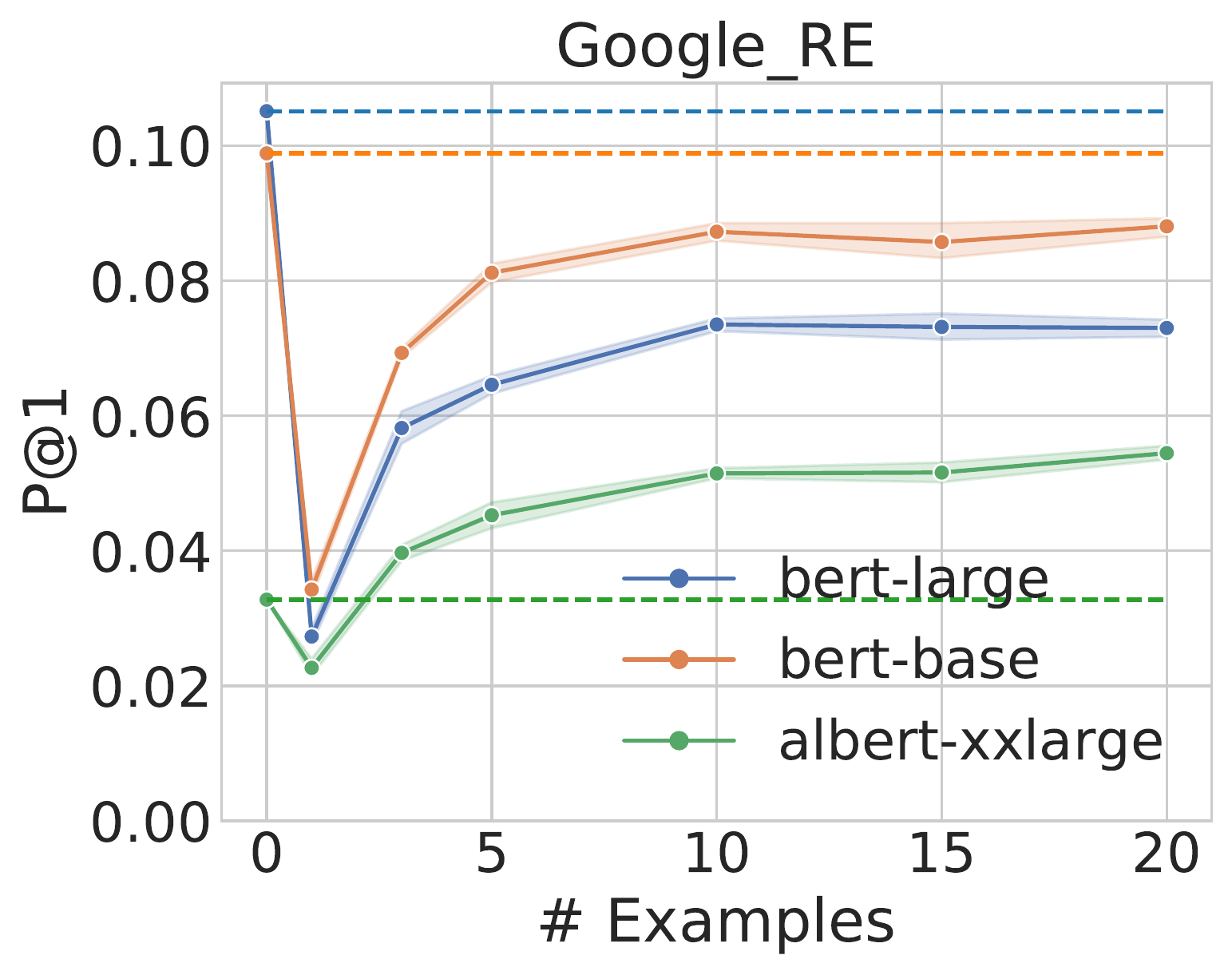}&
    \includegraphics[width=0.25\textwidth]{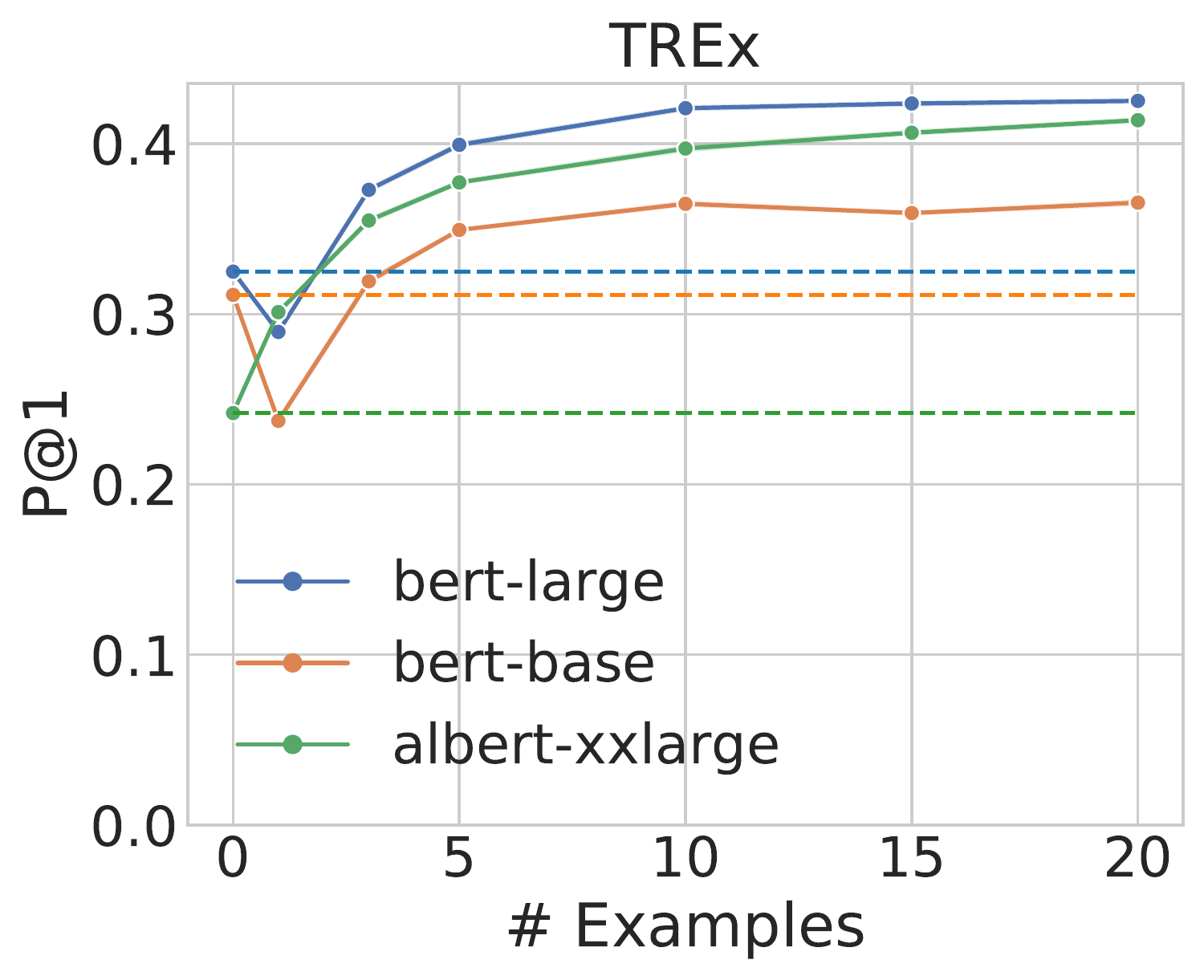}&
    \includegraphics[width=0.25\textwidth]{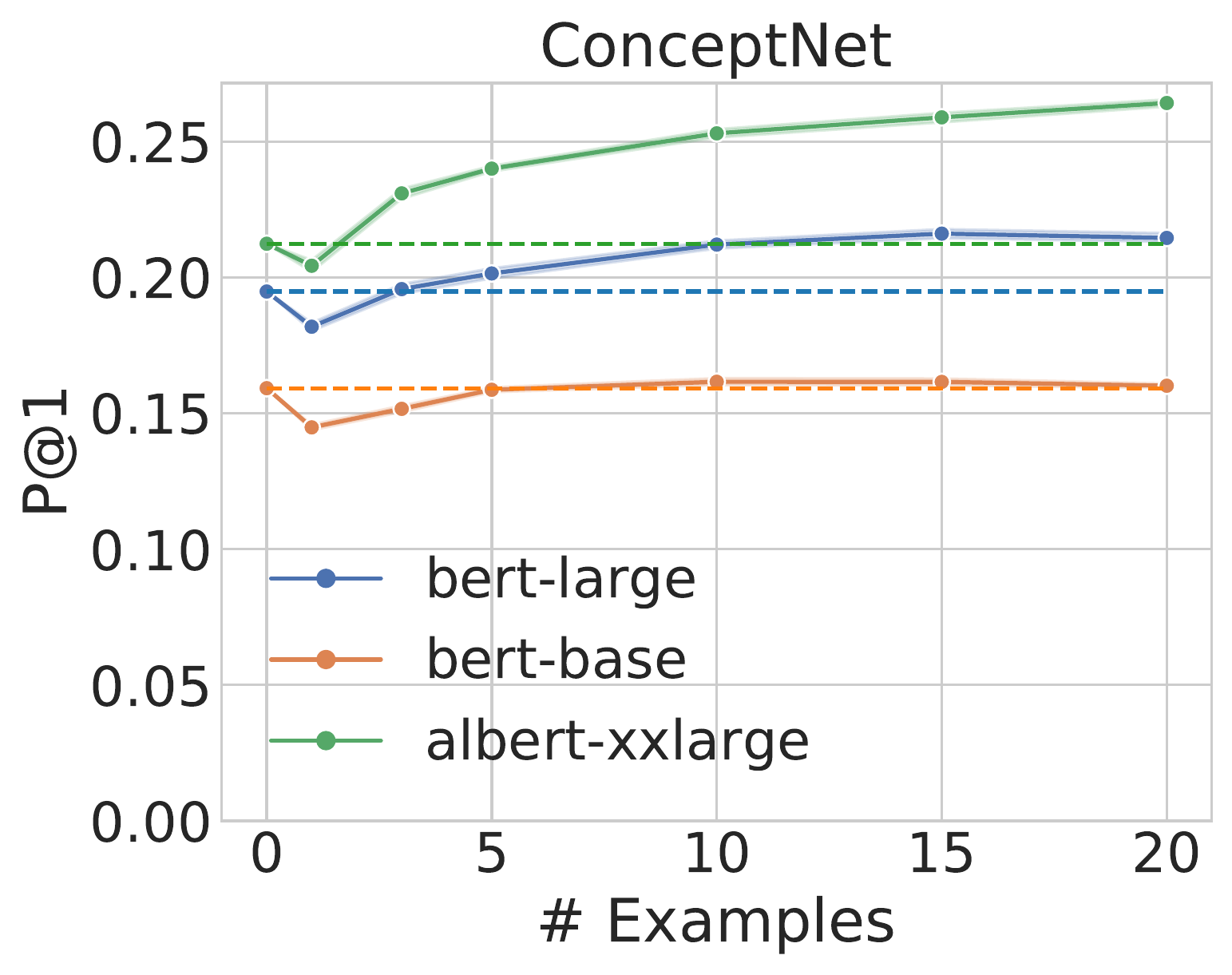} \\
    \hline
    \rotatebox[origin=l]{90}{\hphantom{mmm}Close} &
    \includegraphics[width=0.25\textwidth]{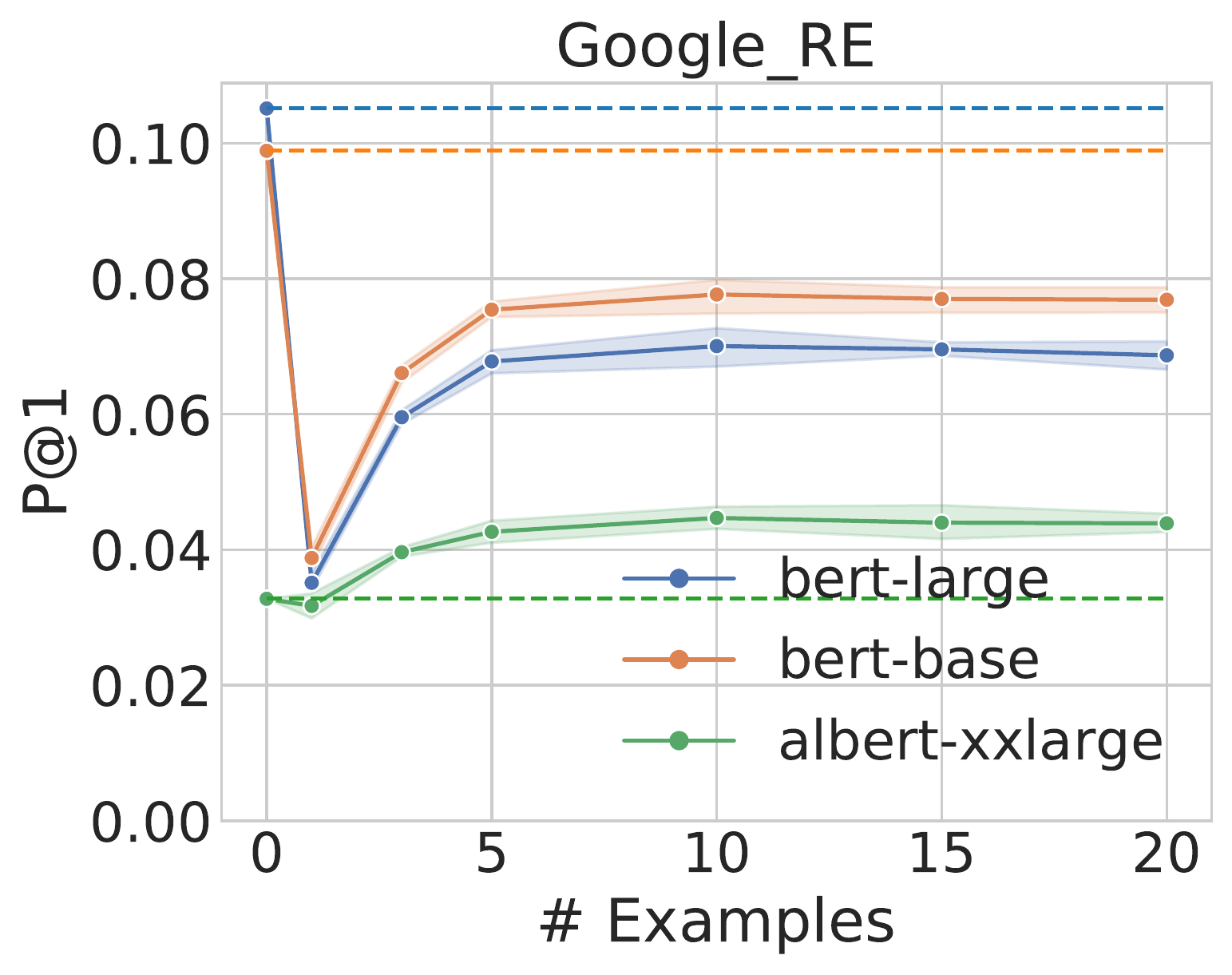}&
    \includegraphics[width=0.25\textwidth]{src/plots/pat1_data-TREx_nl-True_ce-True.pdf}&
    \includegraphics[width=0.25\textwidth]{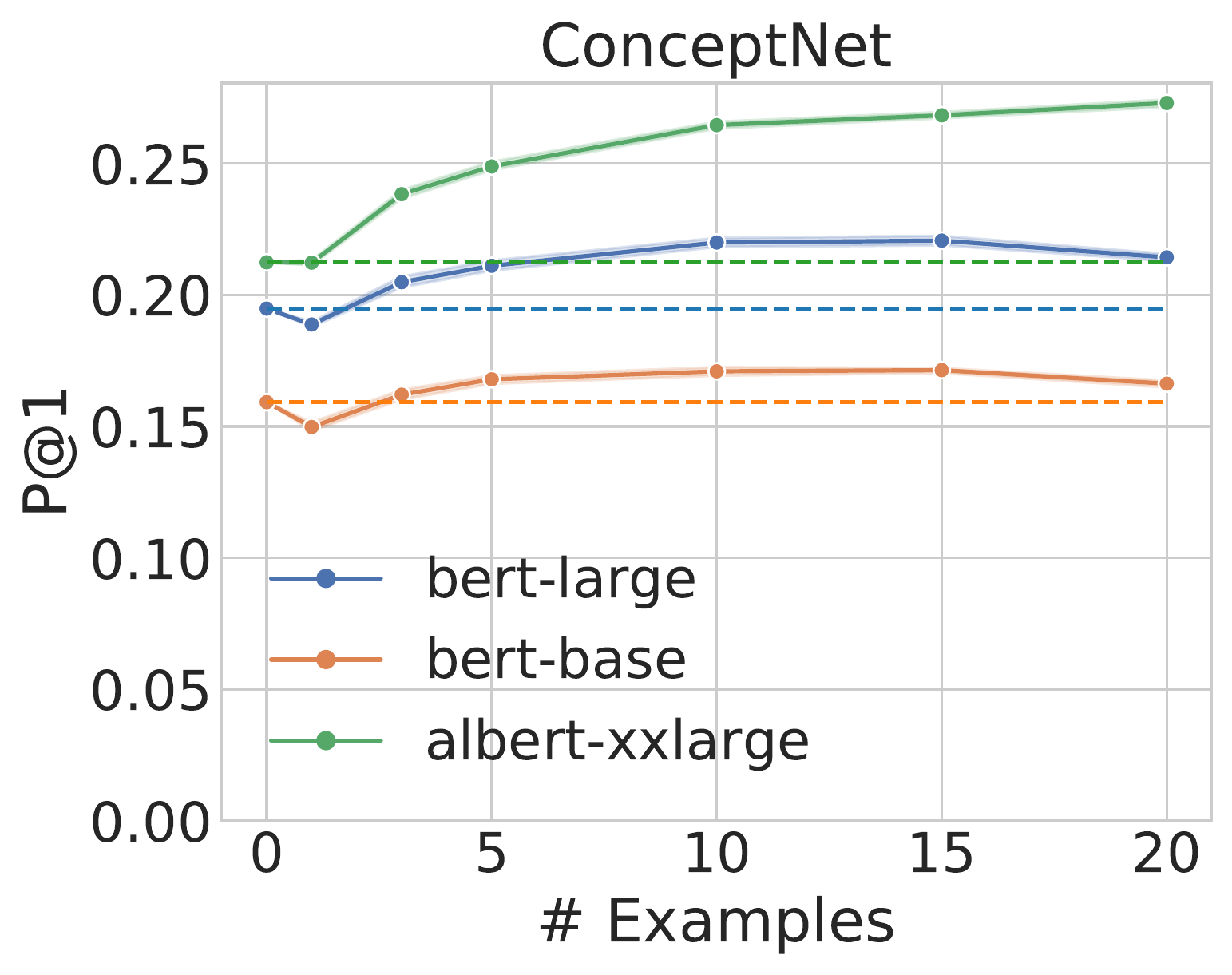} \\
    \bottomrule
    \end{tabular}
}
\caption{P@1 score for the different corpora of the LAMA probe over the number of examples provided. The dashed line shows the baseline values for when no additional example is given. The upper row depicts the scores for when the examples are chosen randomly among the same relation, while the lower row only considers examples from \textit{close} subjects as defined in Section~\ref{sec:method}.}
\label{tab:omitted_figures_p@1}
\end{table}

\begin{table}[h]
\centering
\resizebox{\textwidth}{!}{   
    \begin{tabular}{p{.3cm}c@{}c@{}c@{}}
     & Google-RE & T-REx & ConceptNet \\
    \toprule
    \rotatebox[origin=l]{90}{\hphantom{mmm}Random} & 
    \includegraphics[width=0.25\textwidth]{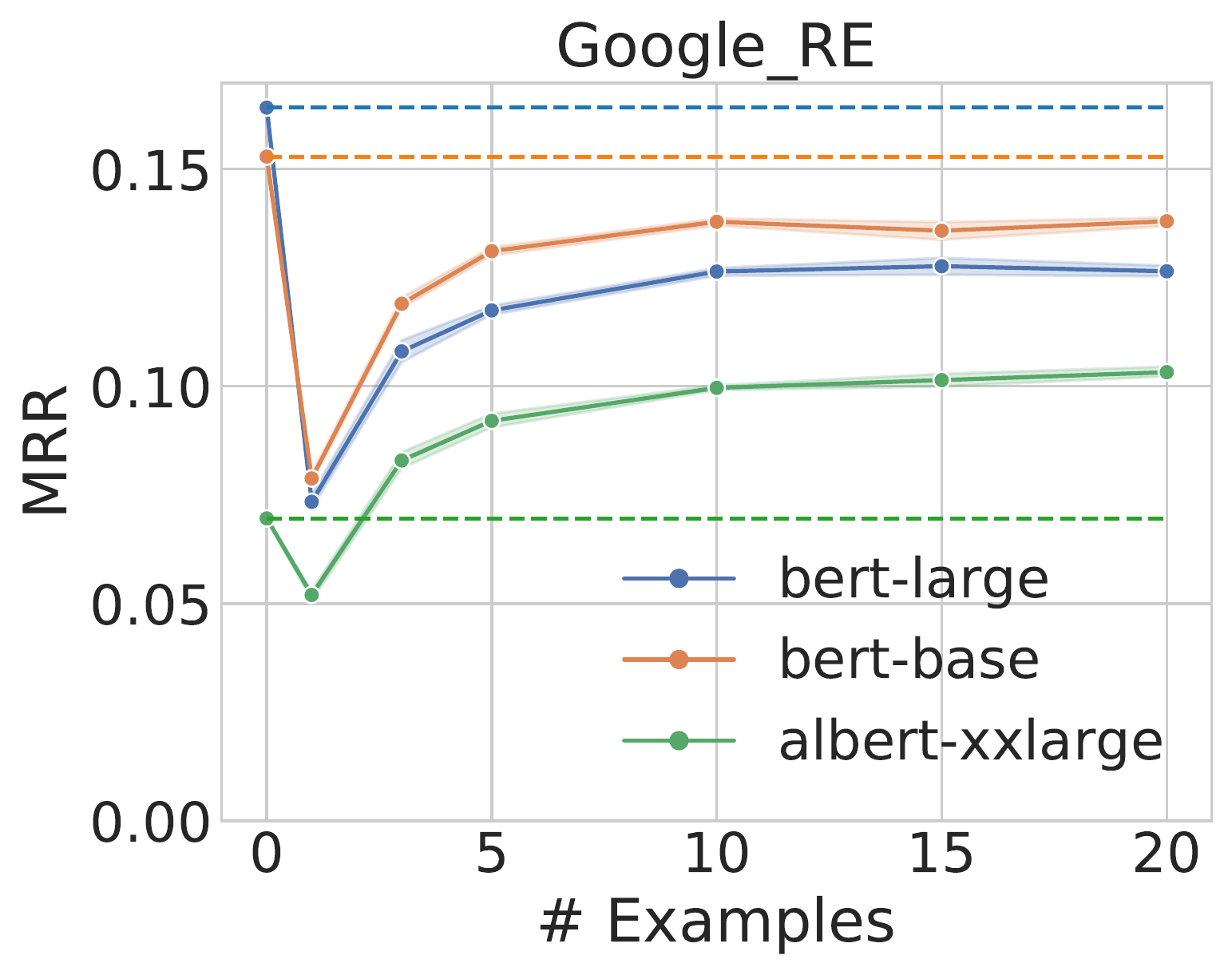}&
    \includegraphics[width=0.25\textwidth]{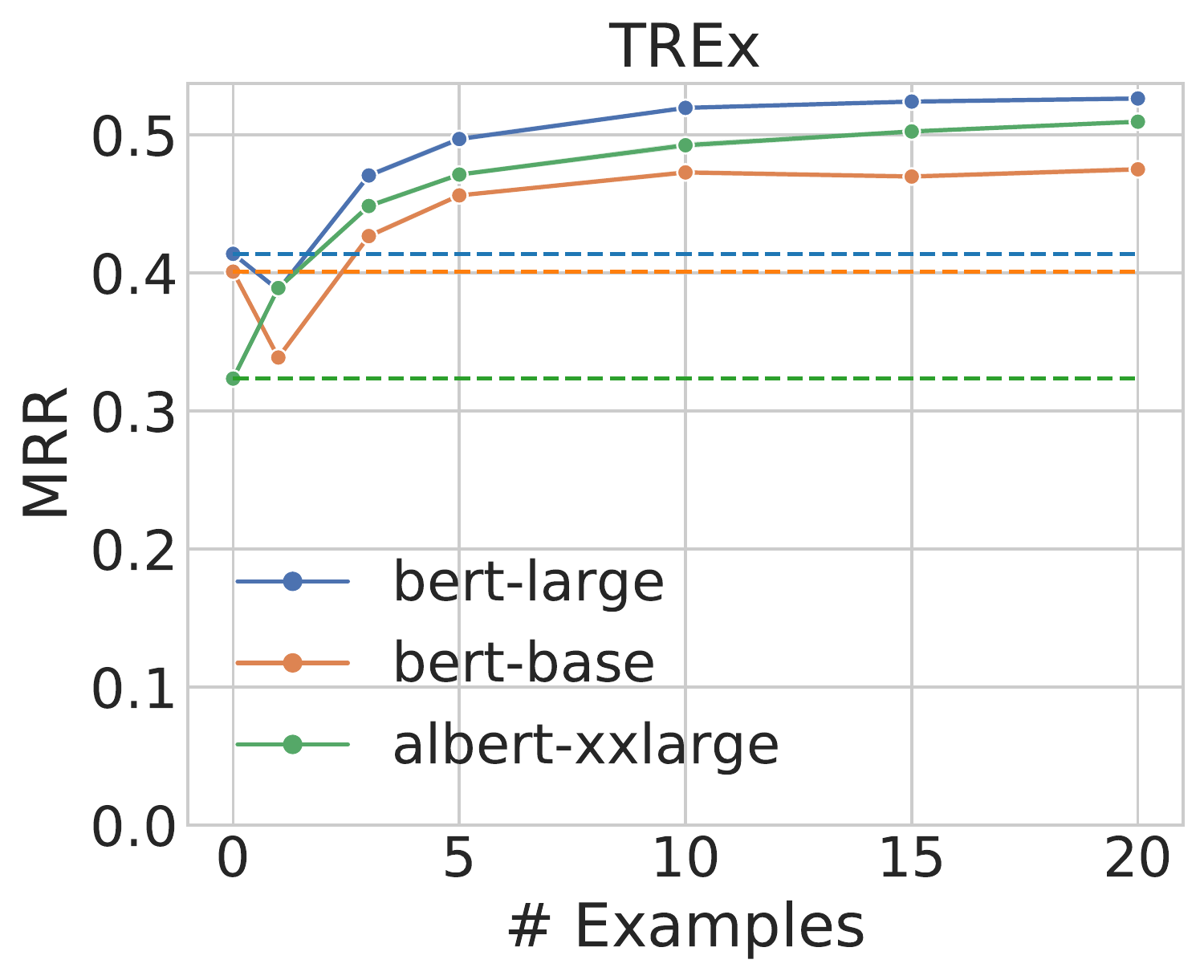}&
    \includegraphics[width=0.25\textwidth]{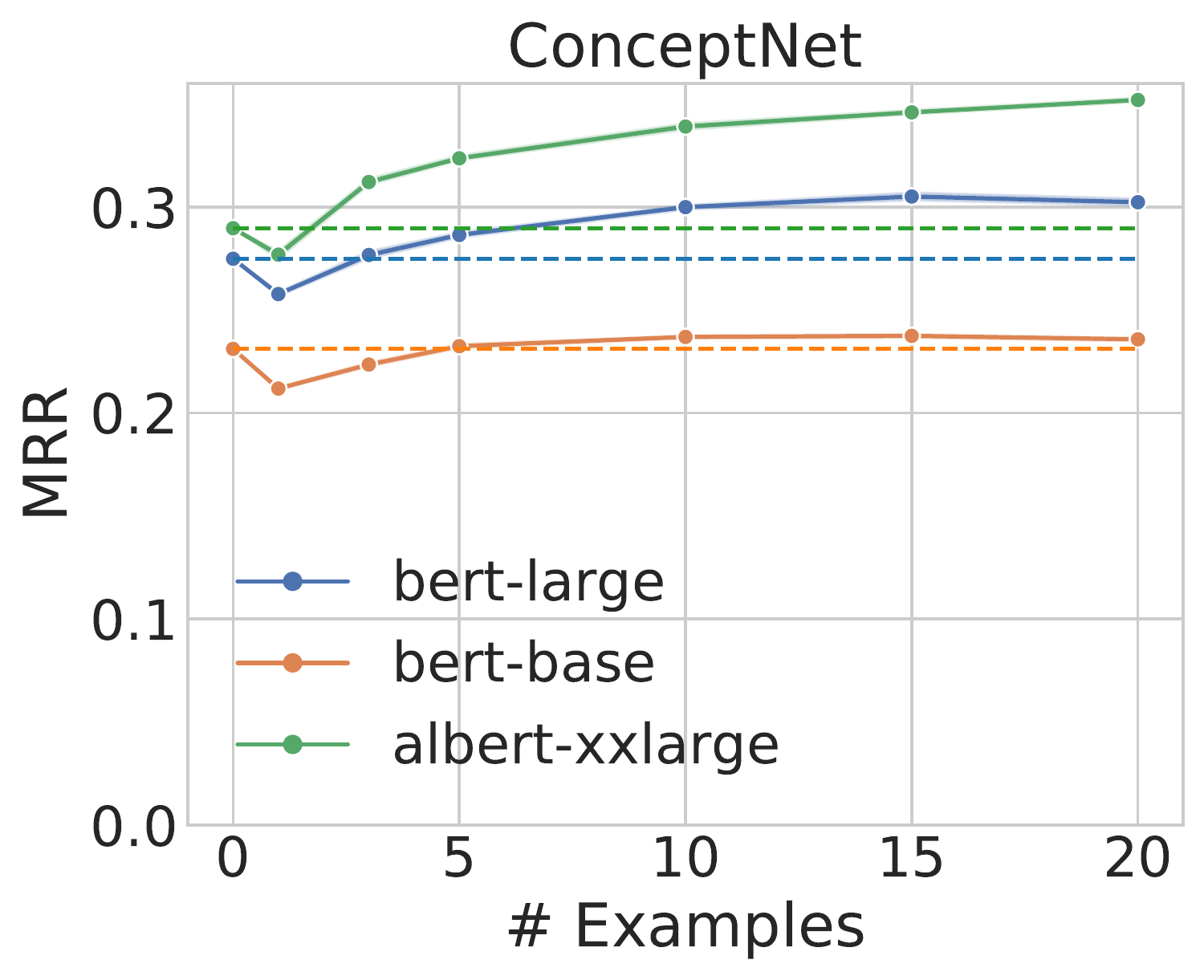} \\
    \hline
    \rotatebox[origin=l]{90}{\hphantom{mmm}Close} &
    \includegraphics[width=0.25\textwidth]{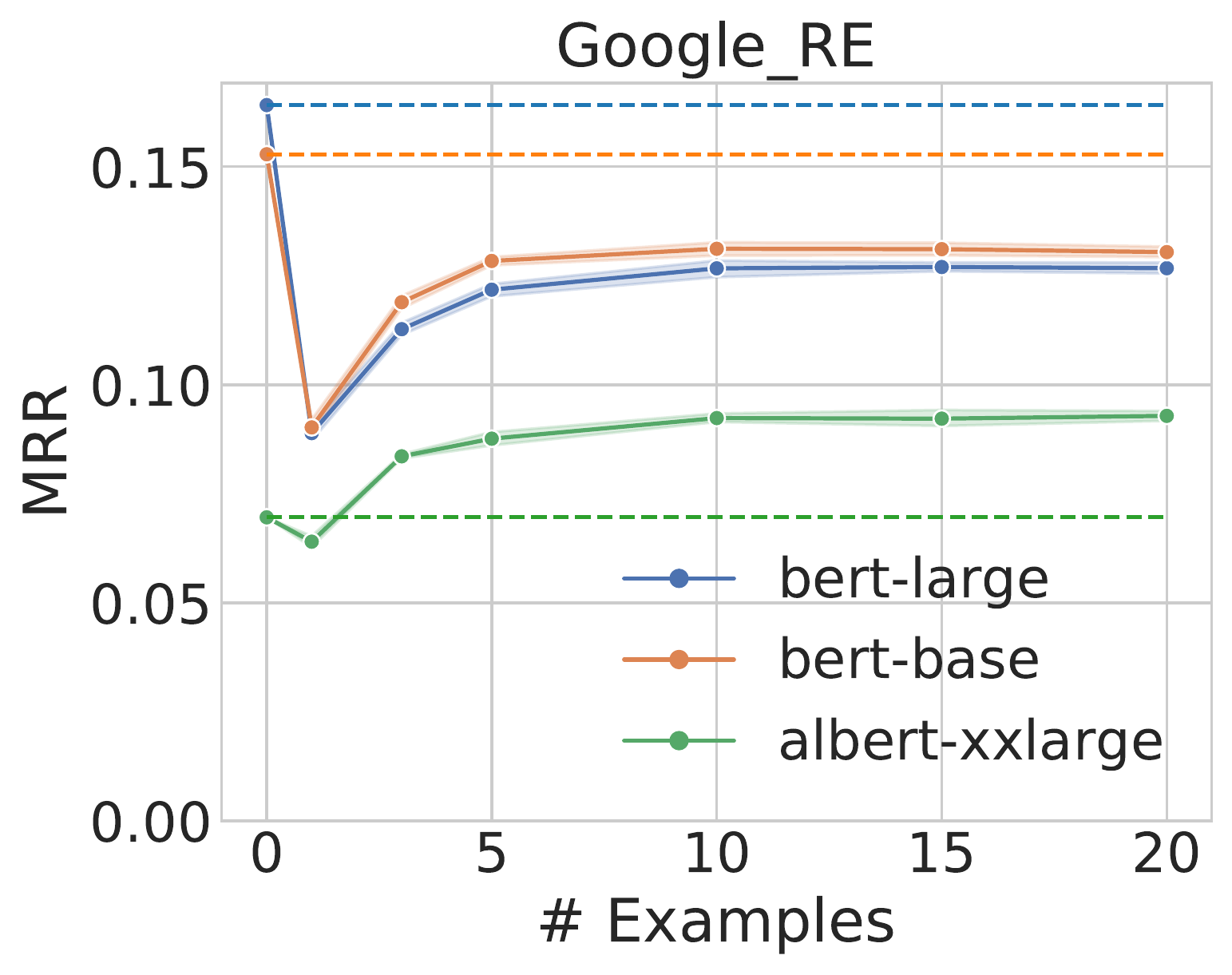}&
    \includegraphics[width=0.25\textwidth]{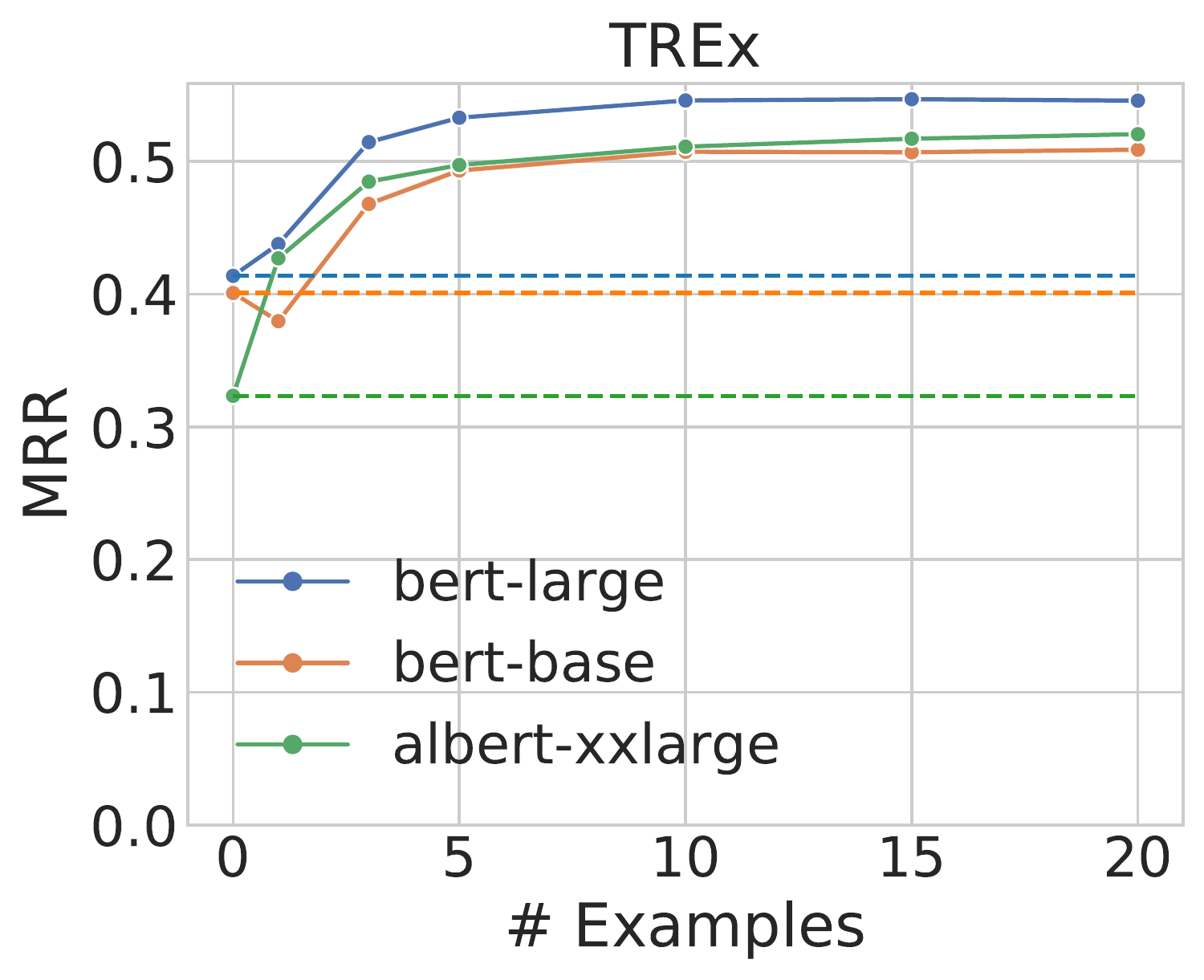}&
    \includegraphics[width=0.25\textwidth]{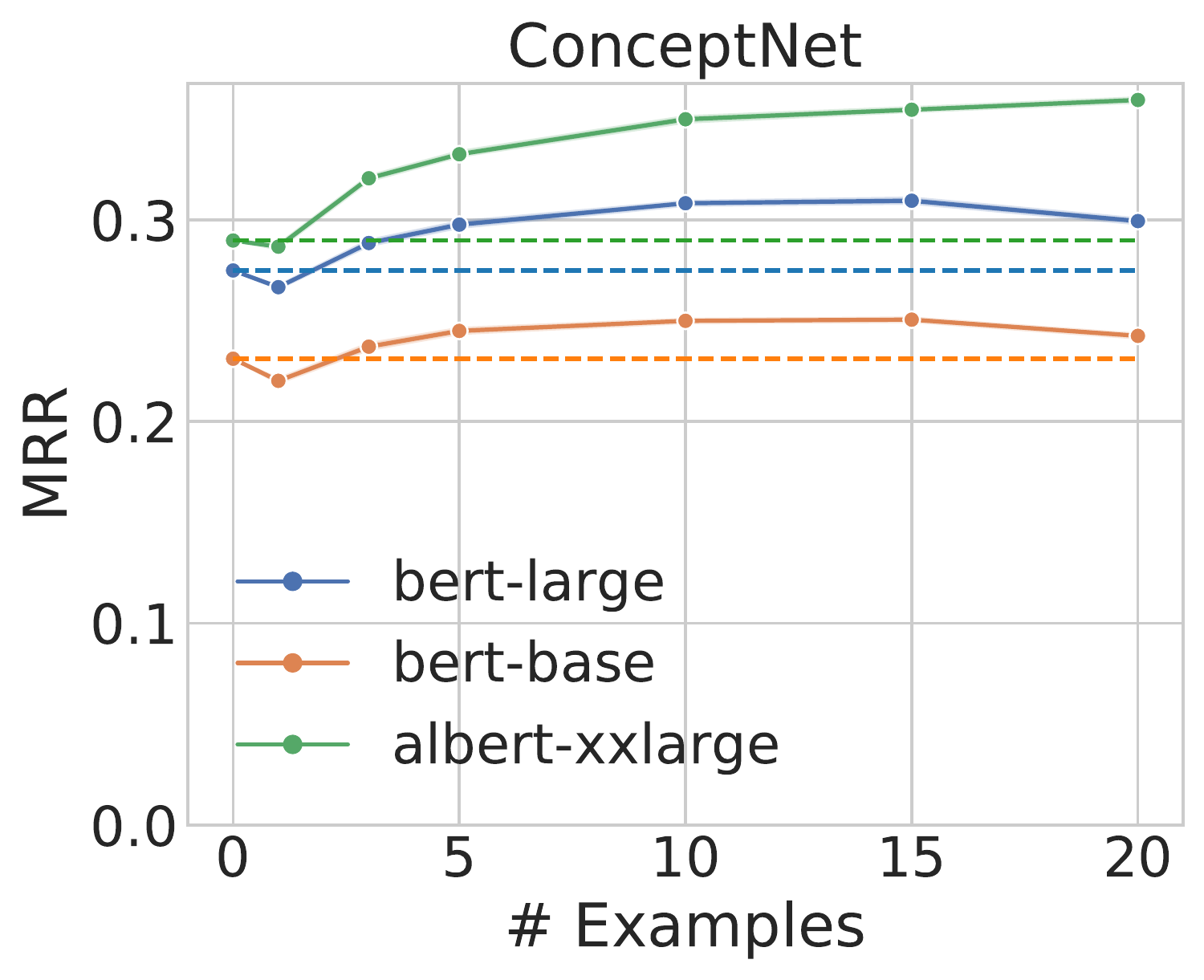} \\
    \bottomrule
    \end{tabular}
}
\caption{Mean reciprocal rank (MRR) score for the different corpora of the LAMA probe over the number of examples provided. The dashed line shows the baseline values for when no additional example is given. The upper row depicts the scores for when the examples are chosen randomly among the same relation, while the lower row only considers examples from \textit{close} subjects as defined in Section~\ref{sec:method}.}
\label{tab:omitted_figures_mrr}
\end{table}

\begin{table}[h]
\centering
\resizebox{\textwidth}{!}{   
    \begin{tabular}{p{.3cm}c@{}c@{}c@{}}
     & Google-RE & T-REx & ConceptNet \\
    \toprule
    \rotatebox[origin=l]{90}{\hphantom{mmm}Random} & 
    \includegraphics[width=0.25\textwidth]{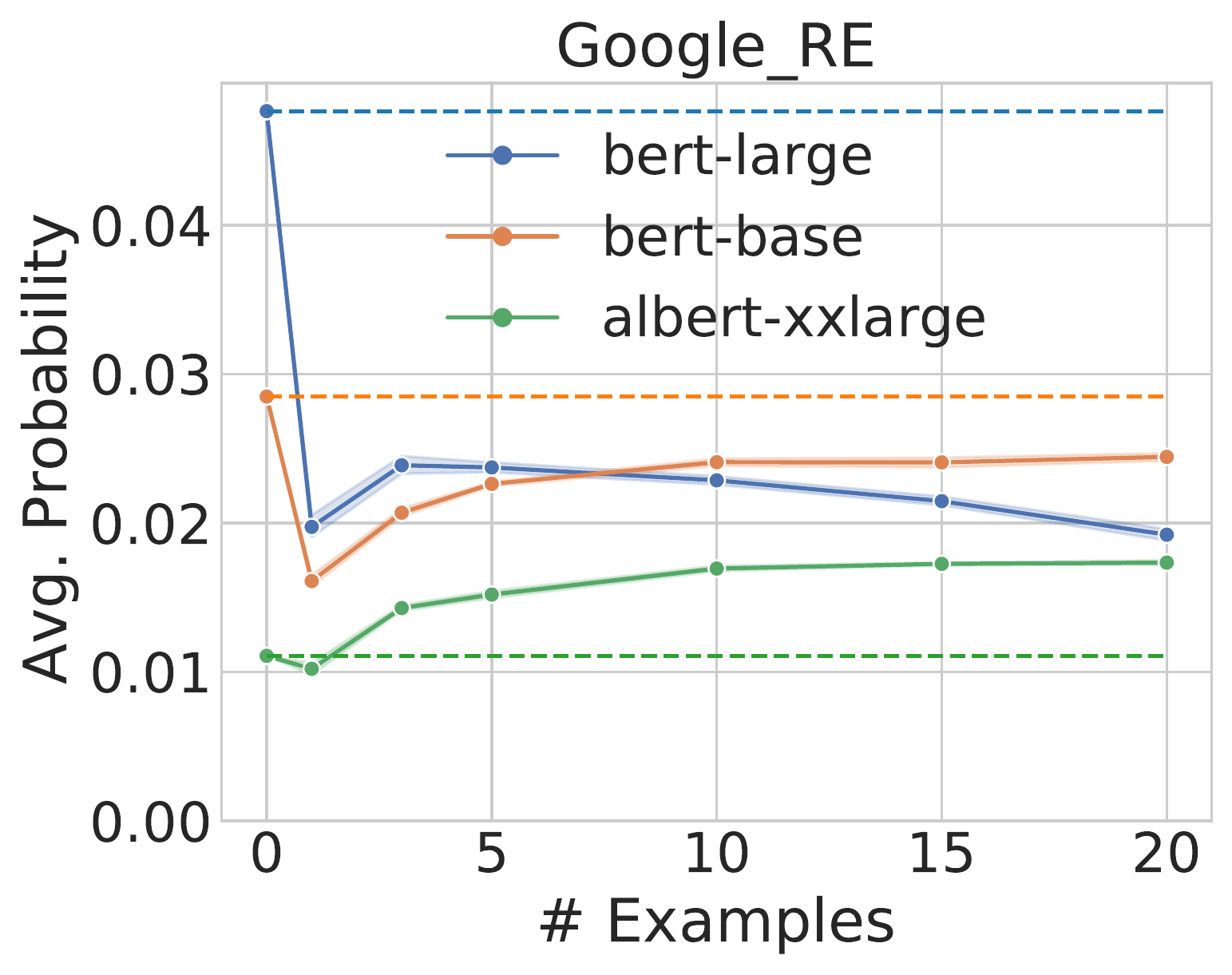}&
    \includegraphics[width=0.25\textwidth]{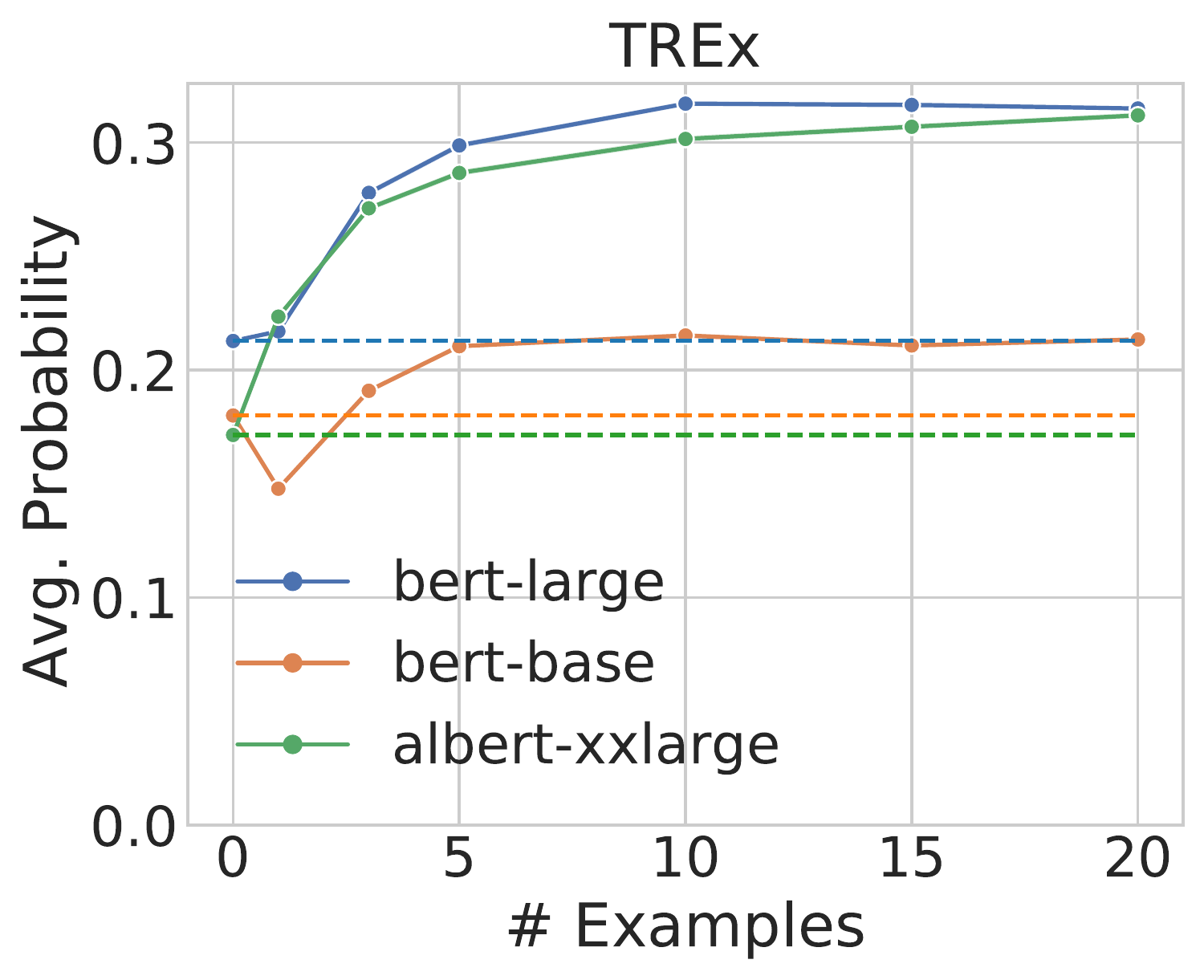}&
    \includegraphics[width=0.25\textwidth]{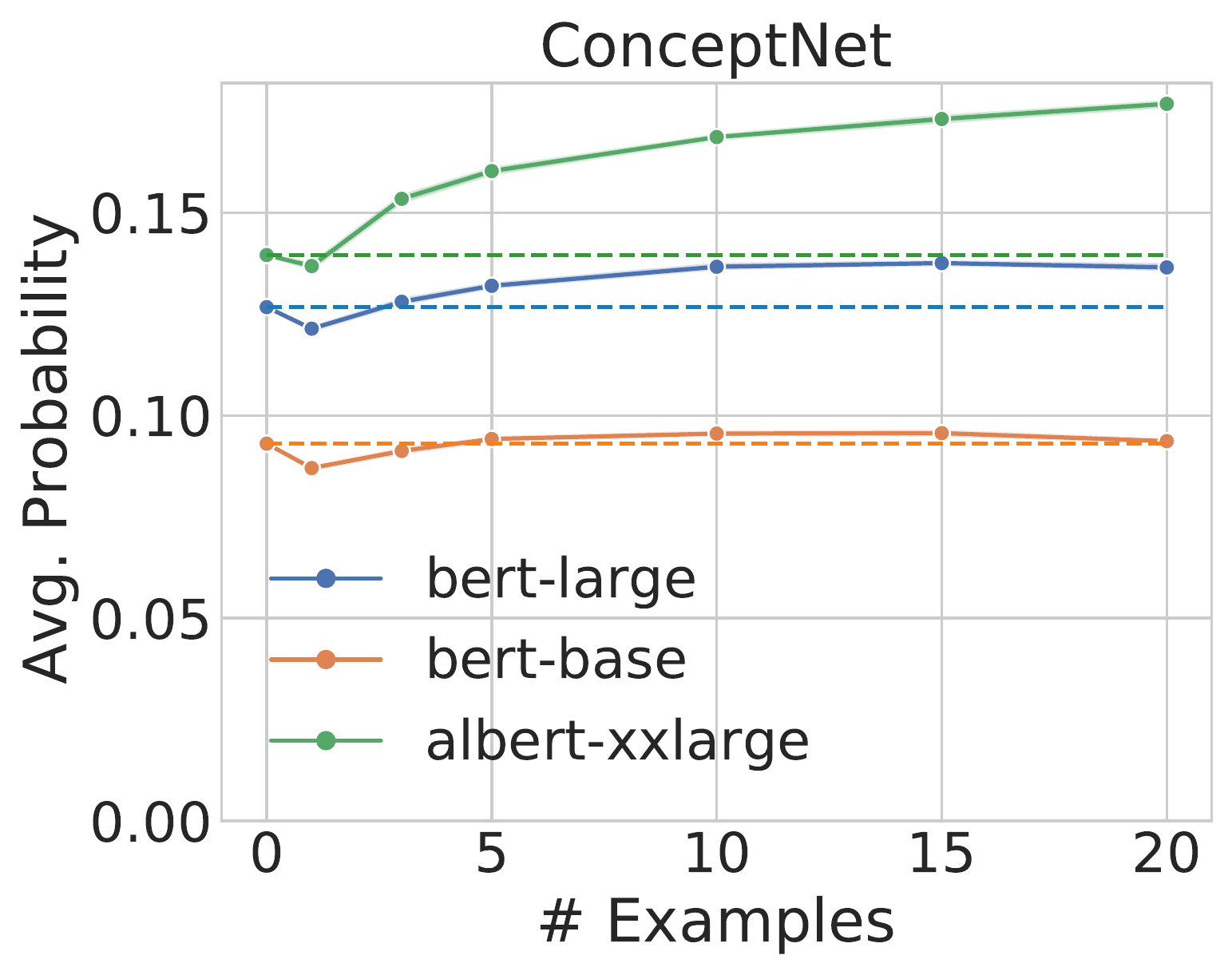} \\
    \hline
    \rotatebox[origin=l]{90}{\hphantom{mmm}Close} &
    \includegraphics[width=0.25\textwidth]{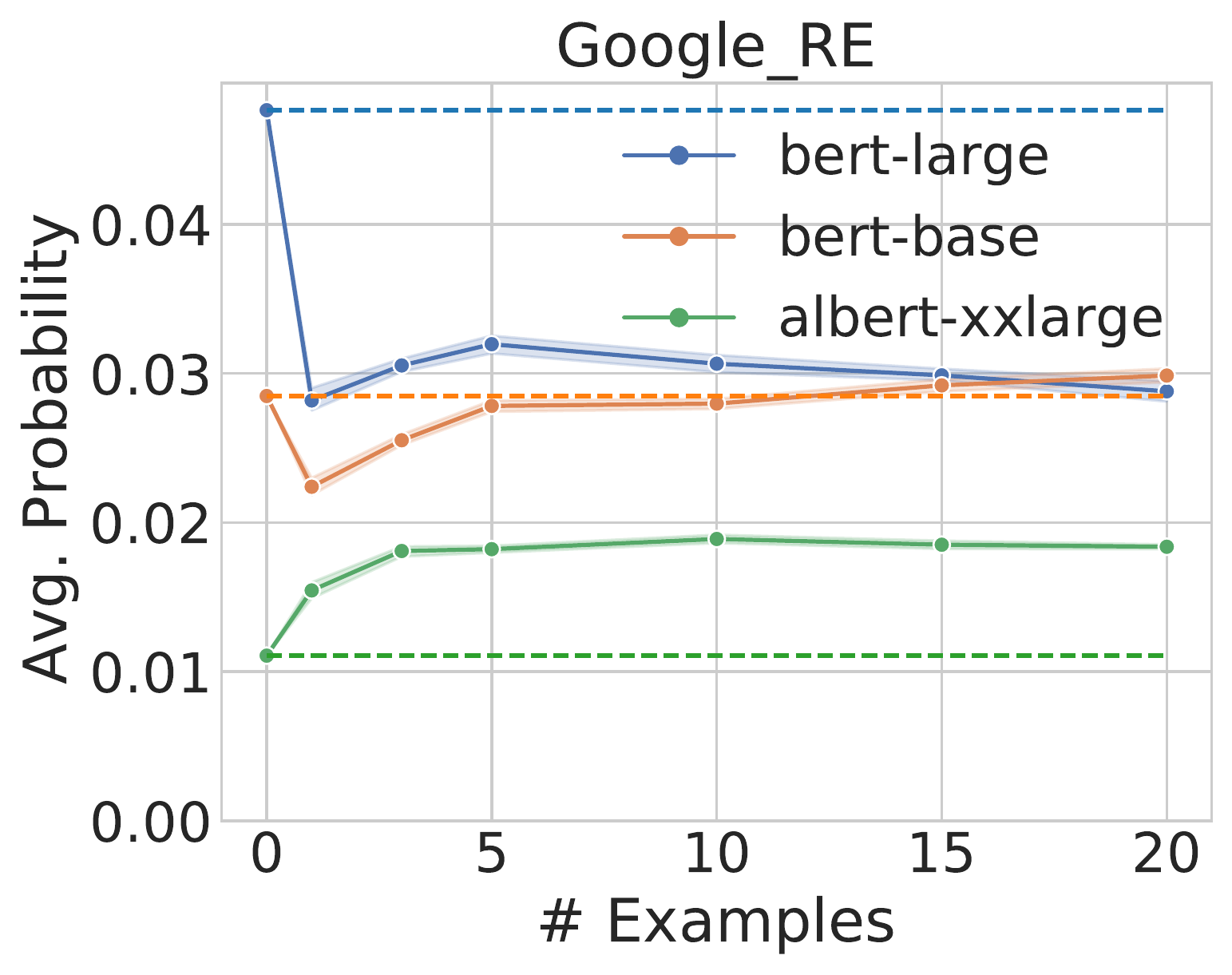}&
    \includegraphics[width=0.25\textwidth]{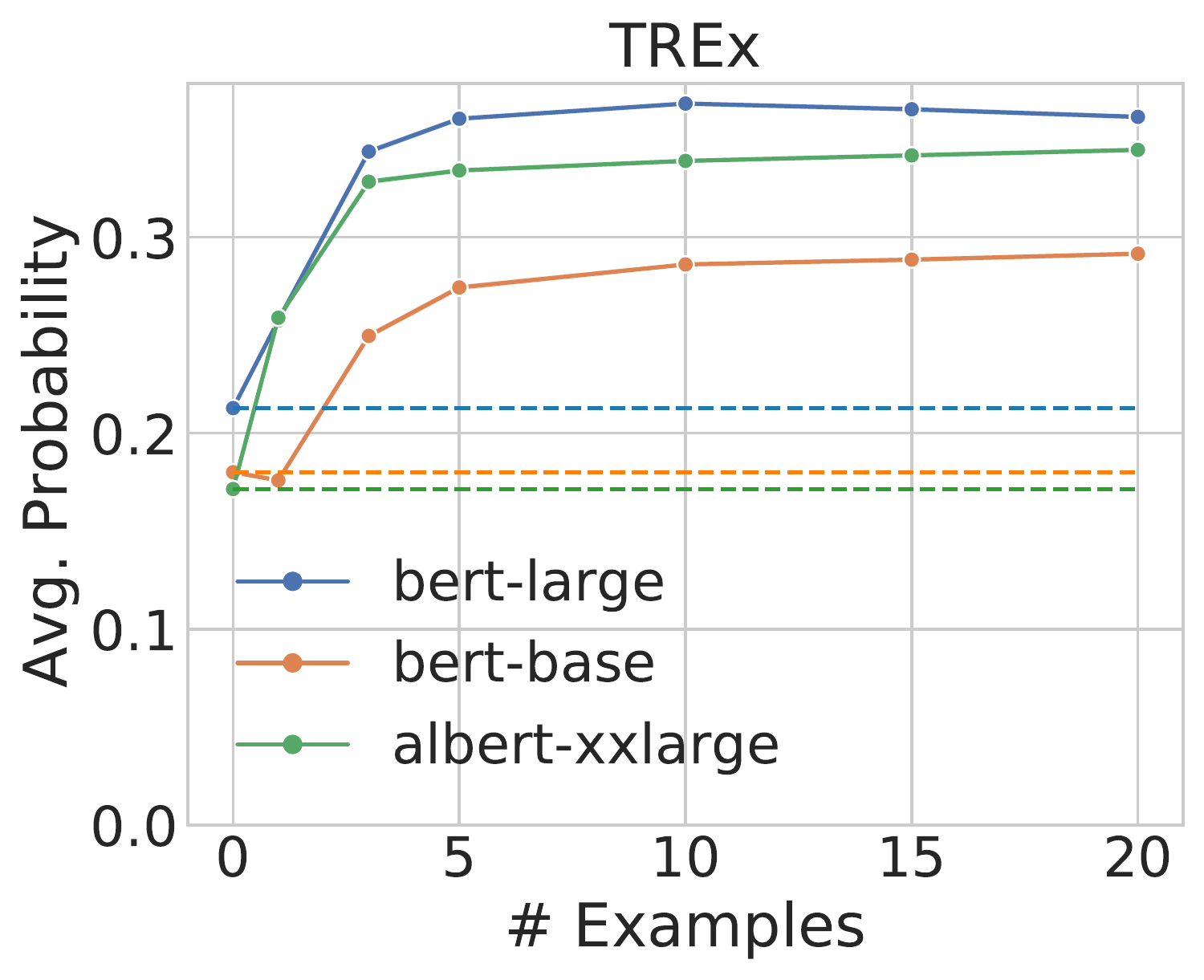}&
    \includegraphics[width=0.25\textwidth]{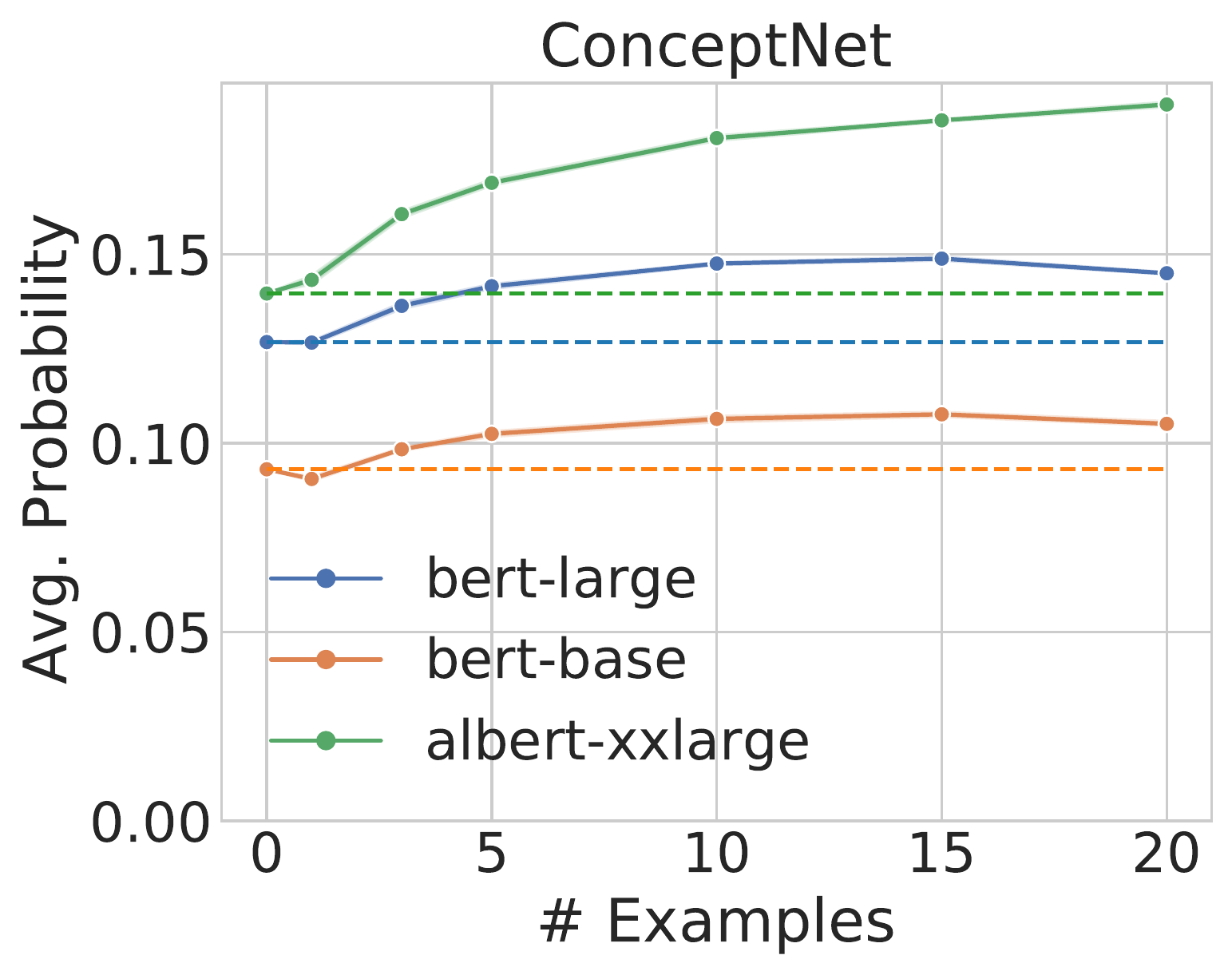} \\
    \bottomrule
    \end{tabular}
}
\caption{Probability assigned to the ground-truth object for the different corpora of the LAMA probe over the number of examples provided. The dashed line shows the baseline values for when no additional example is given. The upper row depicts the scores for when the examples are chosen randomly among the same relation, while the lower row only considers examples from \textit{close} subjects as defined in Section~\ref{sec:method}.}
\label{tab:omitted_figures_target_likelihood}
\end{table}

\begin{figure*}[h]
    \centering
    \includegraphics[width=\textwidth]{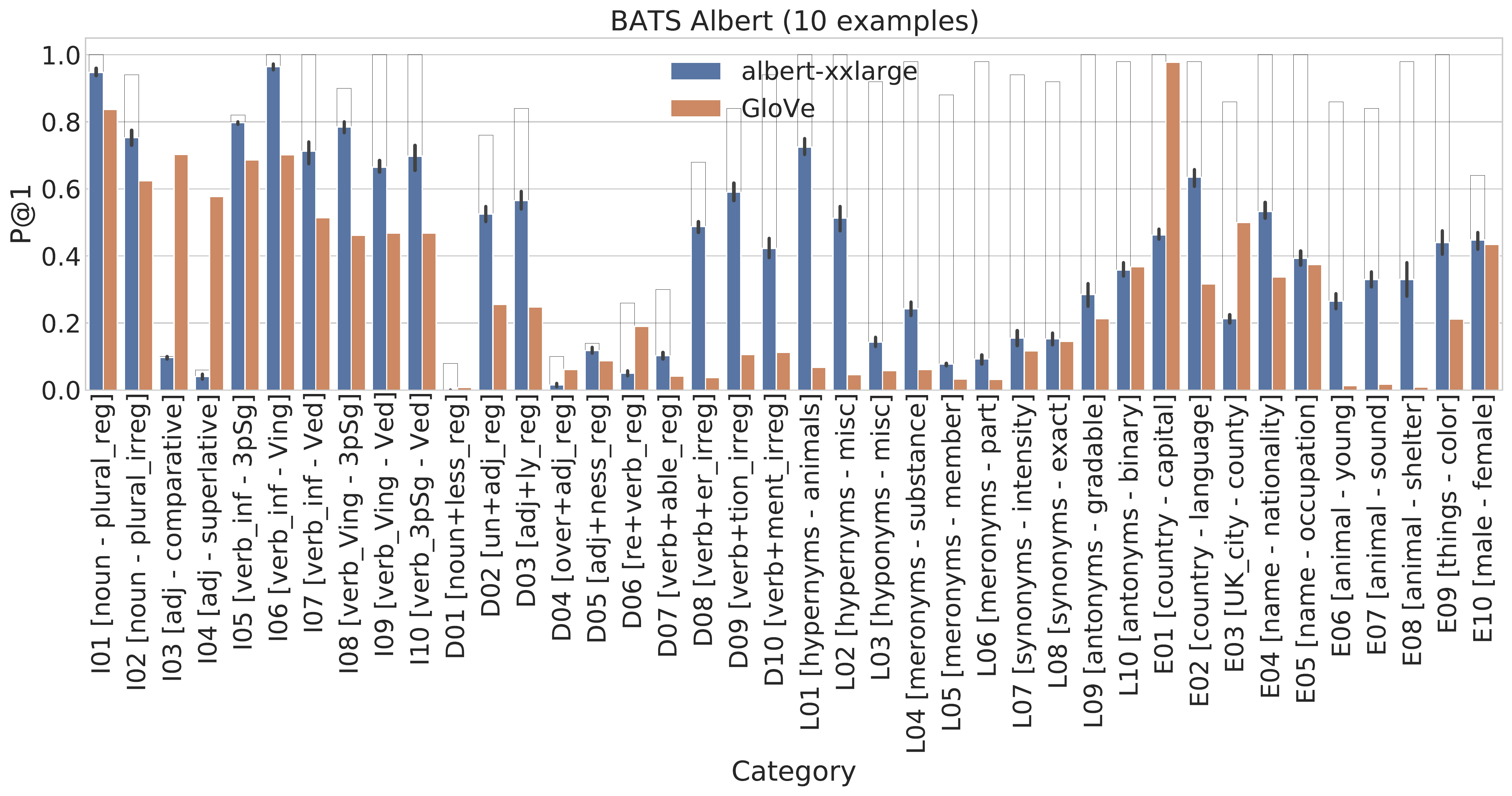}
    \caption{P@1 score on BATS for Albert-xxlarge with 10 examples that use the "([s]; [o])"-template. The x-axis breaks down the performance for the individual relations of the BATS dataset. As a benchmark, we use the GloVe model from \citet{GladkovaDrozd2016}. The frame around the bar indicates the maximum possible score that the Albert model could have scored because not all targets are tokens in its vocabulary.}
    \label{fig:bats_category}
\end{figure*}

\end{document}